# J-EDI QA: Benchmark for deep-sea organism-specific multimodal LLM

Takero Yoshida[1*]   Yuikazu Ito[2]   Yoshihiro Fujiwara[2]   Shinji Tsuchida[2]   Daisuke Sugiyama[1]   Daisuke Matsuoka[1]

[1] Japan Agency for Marine-Earth Science and Technology, Research Institute for Value-Added-Information Generation, Center for Earth Information Science and Technology

[2] Japan Agency for Marine-Earth Science and Technology, Research Institute for Global Change, Marine Biodiversity and Environmental Assessment Research Center

[*] yoshidat@jamstec.go.jp

## Abstract

Japan Agency for Marine-Earth Science and Technology (JAMSTEC) has made available the JAMSTEC Earth Deep-sea Image (J-EDI), a deep-sea video and image archive (https://www.godac.jamstec.go.jp/jedi/e/index.html). This archive serves as a valuable resource for researchers and scholars interested in deep-sea imagery. The dataset comprises images and videos of deep-sea phenomena, predominantly of marine organisms, but also of the seafloor and physical processes. In this study, we propose J-EDI QA, a benchmark for understanding images of deep-sea organisms using a multimodal large language model (LLM). The benchmark is comprised of 100 images, accompanied by questions and answers with four options by JAMSTEC researchers for each image. The QA pairs are provided in Japanese, and the benchmark assesses the ability to understand deep-sea species in Japanese. In the evaluation presented in this paper, OpenAI o1 achieved a 50% correct response rate. This result indicates that even with the capabilities of state-of-the-art models as of December 2024, deep-sea species comprehension is not yet at an expert level. Further advances in deep-sea species-specific LLMs are therefore required.

## 1 Introduction

Multimodal LLMs, which integrate and process not only text but also image, video, and audio data, continue to evolve, and existing models, such as Chat GPT, have demonstrated outstanding performance in multimodal capabilities. A variety of benchmarks have been developed for the purpose of measuring the comprehensibility, versatility, and effectiveness of LLM models. For instance, Gemini [1] employs benchmarks such as MMMU [2], which assesses image comprehension at the college level, and MathVista [3], which gauges mathematical comprehension from images. Similarly, LLaVA [4] and VILA [5], which are open-source multimodal LLMs, have published evaluation results using benchmarks such as TextVQA [6], MM-Vet [7], SEED-Bench [8], MME [9], and GQA [10]. There are additional image-related benchmarks (e.g., [11-14]). Furthermore, the MMMU mentioned above has been updated to MMMU-Pro [15] and continues to evolve in line with the development of LLM.

Other benchmarks, such as VATEX [16] and Perception Test [17], have been employed for the purpose of evaluating video understanding. In order to achieve AGI, there are additional benchmarks, which focuses on everyday activities and interactions. While existing datasets and benchmarks, such as Perception Test, which assesses comprehension of general videos, are important, the construction of domain-specific LLMs and benchmarks will facilitate the use of LLMs in many fields including science in the future. With a focus on scientific use cases, for instance, the Science QA [18] benchmark exists for the purpose of understanding and analyzing scientific images. Meanwhile, MMMU is said to be a college-level problem, however it is desirable to include content that extends further up to the expert level.

In light of the above, the J-EDI QA presented in this paper represents a distinctive benchmark for the analysis of image content obtained from deep-sea surveys. This benchmark differs from others in that it focuses on the specific domain of deep-sea imagery, namely the recognition of deep-sea organisms, rather than assessing a broad range of routine image recognition capabilities. The details of the J-EDI QA, a benchmark for evaluating the image understanding capability of multimodal LLMs specific to deep-sea organisms, are presented in this paper. This benchmark is also composed in Japanese, as exemplified by Heron-Bench [19], JA-VLM-Bench-In-the-Wild [20], and JMMMU [21] which are a Japanese-specific image benchmark designed to measure the comprehension ability for Japanese terminology.

The J-EDI dataset comprises a diverse array of deep-sea images of deep-sea organisms. From this corpus, 100 images were selected by JAMSTEC researchers and accompanied by QA pairs designed to assess the ability to identify species. Each QA pair consists of four multiple-choice questions and answers presented in Japanese, thereby enabling the measurement of both deep-sea organisms and Japanese language comprehension.

This paper shows the results of using this benchmark to evaluate the solution performance of the Chat GPT. The findings indicate a need for the development of additional scientific and expert-level benchmarks for scientific domain-specific LLM in the future.

In the future, the development of multimodal LLM models specific to deep-sea organisms will streamline the process of identifying organisms and their distribution status from deep-sea imagery. It will also lead to the development of deep-sea probes equipped with a function that automatically identifies organisms. In addition, although this benchmark is being created for the purpose of evaluating multimodal LLM, it can also be used as a test to measure the public's understanding of deep-sea organisms. This can also be used as a deep-sea education and outreach activity.

# 2 Dataset

## 2.1 J-EDI

J-EDI, a deep-sea video and image archive developed by JAMSTEC's Global Oceanographic Data Center (GODAC), is accessible via the Internet. In addition to the videos and images, the database includes information such as the location where the images were taken and the general species of organisms. As of July 2021, the archive comprises images and videos captured between 1982 and 2019. The total number of images released to the public has exceeded one million. The images were primarily captured in the deep waters surrounding Japan. While the majority of the dataset comprises marine organisms, other types of deep-sea content are also included, such as seafloor rocks and physical phenomena, thereby providing a diverse range of deep-sea-related content. Additionally, the dataset includes rare deep-sea species. Furthermore, phenomena that are crucial for understanding deep-sea ecosystems, such as undersea volcanic activity and hydrothermal vent ecosystems, are also represented.

100 images were selected by JAMSTEC researchers and accompanied by QA pairs. The data size of each of the 100 images used as this benchmark is approximately 0.5 to 3 megabytes, which allows for the recognition of organisms while maintaining sufficient image quality. Each content item can be downloaded directly from the web page, but prior to downloading, users must agree to the following terms of use and indicate the purpose for which the content will be used.

J-EDI Terms of Using Videos and Photos

The copyright of videos and photos of this site belongs to JAMSTEC.
By downloading videos and/or photos from this site, you agree to the following terms and conditions.
1. JAMSTEC shall not be responsible for any loss, damages and troubles you make either directly or indirectly by using downloaded videos and/or photos of this site.
2. You are not permitted to manipulate, redistribute, deliver, or sell downloaded videos and/or photos of this site to public.

3. Although using data for nonprofit scientific or educational purposes is free of charge, you are requested to express that JAMSTEC data is used in your paper, document or video, etc.
4. If you use videos and photos of this site for non-educational or non-scientific purposes, such as commercial purpose, etc., or if you need to process or edit them, please inform us in advance and follow our instructions.

## 2.2 QA

In the QA phase, the identification of identifiable species of organisms was conducted by researchers from JAMSTEC, and a selection question was included to facilitate the potential for erroneous responses.

Chat GPT was tasked with stating the chosen answer and providing a rationale for the selection, as well as outputting the criteria used for evaluation. This information is supplementary to the evaluation of the model's comprehension of the organisms depicted in the images. To facilitate evaluation, we utilized only multiple-choice questions for simplicity. However, by having LLMs respond to questions regarding the identification of organisms from images and their associated judgments, it will be feasible to conduct a manual evaluation of the discrepancies between the experts' explanations. Furthermore, this commentary by experts could be transferred for training purposes when constructing large datasets. In this paper, the dataset consists of only 100 images, which is somewhat small even for a benchmark, nevertheless we are considering increasing the number of images in the future.

The specifics of the benchmarks are outlined in the table at the last part of this paper. Secondary use is subject to the same conditions as the terms of use outlined in the above section, and is limited to non-commercial purposes, such as research and education.

# 3 Evaluation

## 3.1 Evaluation method

The percentage of correct answers to the selection questions was employed as a simplified assessment of the model's accuracy. It represents the proportion of questions answered correctly out of the total number of questions, and serves as a measure for evaluating the overall performance of the model. The evaluation was conducted by responding to multiple-choice questions using OenAI o1 and GPT-4o. An image was uploaded, and the model was presented with a question and four potential answers, along with instructions to select one and provide a rationale. While the species information and general classification are also displayed on the J-EDI archive, in the benchmark evaluations described below, only images were uploaded, and no information on the species listed on the web page was provided.

The percentage of correct responses obtained from the model was then determined. The prompts are presented below, accompanied by an illustrative example of a response generated by GPT-4o. A total of 100 images were presented with the same prompt. The results were employed to assess the discriminatory capacity of the existing models in relation to the deep-sea organism-specific benchmarks presented in this paper, based on their comprehension and familiarity with the Japanese terminology.

Example 1
Prompt：
"画像を見て、以下の問題に答えてください。問題は選択式です。また答えについての解説もお願いします。画像の生物は何でしょうか？ A.オンデンザメ B.エドアブラザメ C.カグラザメ D.シロカグラ (Please direct your attention to the provided image and respond to the following inquiries. Please note that these questions are optional. In addition, please provide a rationale for your response. What is the identity of the organism depicted in the image? A. *Somniosus pacificus* B. *Heptranchias perlo* C. *Hexanchus griseus* D. *Hexanchus nakamurai*)"

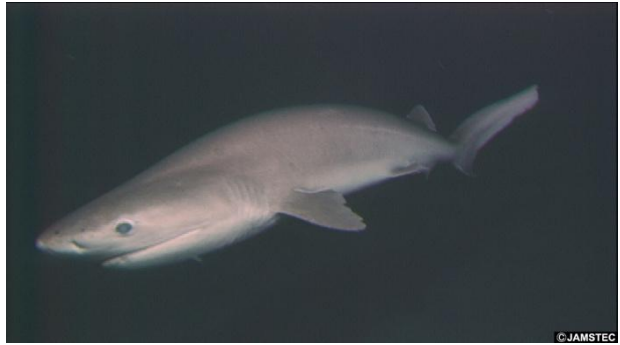

Figure 1: Example image, referenced from https://www.godac.jamstec.go.jp/jedi/player/j/HPD1014HDTV0126

Answer by GPT-4o：

"この画像の生物は「エドアブラザメ」です。(The creature in this image is the *Heptranchias perlo*.)
解説(explanation):

　　画像に写っているサメは、典型的なエドアブラザメの特徴を持っています。エドアブラザメは深海に生息するサメの一種で、体が比較的丸みを帯びており、頭部が大きく、鋭い目が特徴です。背びれは低く、全体的に滑らかな外観をしています。(The shark depicted in the image exhibits the typical characteristics of the species *Heptranchias perlo*. *Heptranchias perlo* is a deep-water dwelling shark that is distinguished by a relatively rounded body, a large head, and sharp eyes. The dorsal fin is low and has an overall smooth appearance.)"

Expert Answers and Commentary：

"カグラザメ(*Hexanchus griseus*)"

"大型で鰓孔が 6 つあり、体は灰色、背びれが 1 基で体の後方にある。(The species is characterized by a large size, six gill slits, a gray body color, and a single dorsal fin located behind the body.)"

Example 2

Prompt：

"画像の生物は何でしょうか？ A.アシナガツノガニ B.オーストンガニ C.タカアシガニ D.ノコギリガニ (What is the identity of the organism depicted in the image? A. *Phalangipus* hystrix B. *Cyrtomaia* curviceros C. *Macrocheira kaempferi* D. *Schizophrys aspera* ")"

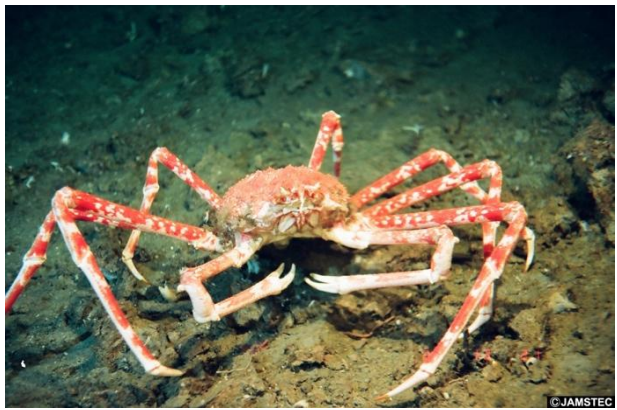

Figure 2: Example image, referenced from https://www.godac.jamstec.go.jp/jedi/player/j/2K0709IN0062Hp02-25

Answer by GPT-4o：

"この画像の生物は「タカアシガニ」です。(The organism in this image is the *Macrocheira kaempferi*.)

Explanation:

 画像に写っているカニは、非常に長い脚を持っており、甲羅が小さく脚が非常に発達しているのが特徴です。これらの特徴は、世界最大のカニであるタカアシガニに典型的です。(The crab depicted in the image exhibits notably elongated legs, a diminutive carapace, and highly advanced limb development. These traits align with those observed in the *Macrocheira kaempferi*, a species renowned for its immense size, which is regarded as the largest crab in the world.)"

Expert Answers and Commentary：

"タカアシガニ（*Macrocheira kaempferi*)"

"非常に大型で、脚が非常に長い。体色は赤く、脚には白色のまだら模様が入る。(The specimen is notably large, with exceptionally elongated legs. The body is red in color, while the legs display white mottled patterns.)"

Example 3

Prompt：

"画像の生物は何でしょうか？ A.シンカイハゲナマコ B.ウロコムシ C.ワタゾコナマコ D.センジュナマコ (What is the identity of the organism depicted in the image? A. *Pannychia henrici* B. Polynoidid worm C. *Benthodytes* sp. D. *Scotoplanes globosa* )"

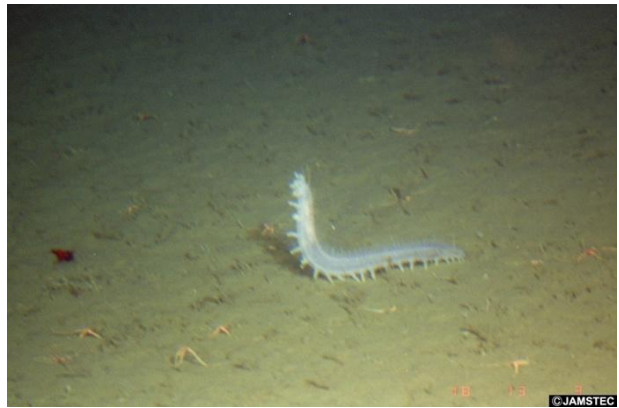

Figure 3: Example image, referenced from https://www.godac.jamstec.go.jp/jedi/player/j/2K1129IN0099Hp04-22

Answer by GPT-4o：

"この画像の生物は「ウロコムシ」です。(The organism in this image is the Polynoidid worm.

Explanation:

 ウロコムシは、多毛類に属する海洋生物で、体全体が鱗状の構造で覆われているのが特徴です。この画像では、砂地に立ち上がった姿が見られ、体側から伸びる付属肢が確認できます。この付属肢がウロコムシの特徴的な形状を作り出しています。(The Polynoidid worm is a marine organism belonging to the polychaete group, distinguished by a scaly covering of the body. In this image, the organism is observed in an upright position on a sandy surface, with appendages extending from the lateral surfaces of its body. These appendages contribute to the distinctive morphology of the scale worm.)"

Expert Answers and Commentary：

"シンカイハゲナマコ（*Pannychia henrici*)"

"体は円筒形で長い。体色は白～乳白色～淡紫褐色。腹面の両側にそれぞれ管足列がある。背側に 2 列の小さ

な疣足とその間に 8 対くらいの長い鬚状の疣足がある。(The body is cylindrical and elongated. The coloration of the body ranges from white to milky white to light purple-brown. There are rows of tube feet on both sides of the ventral surface. On the dorsal side, there are two rows of small tube feet, with approximately eight pairs of long whisker-like tube feet in between.)"

### 3.2 Evaluation result

The percentage of correct responses on the GPT-4o is 39% (39 out of 100 questions answered correctly). A more detailed analysis of correct answers is as follows:

Fishes: 41.7% (15 of 36 questions answered correctly)

Crustaceans: 45.0% (9 of 20 correct)

Other invertebrates: 34.1% (15 of 44 correct)

Similarly, the percentage of correct responses on the o1 is 50% (50 out of 100 questions answered correctly). A more detailed analysis of correct answers is as follows:

Fishes: 47.2% (17 of 36 questions answered correctly)

Crustaceans: 70.0% (14 of 20 correct)

Other invertebrates: 43.2% (19 of 44 correct)

Comparing the results, o1 had a higher percentage of correct answers than GPT-4o, correctly selecting about half of the questions. Correct answers were particularly high for crustacean questions, probably because crustaceans have more distinctive features than fish and other invertebrates. Some images included pictures where it was difficult to identify the organism from the images given alone, as some of the features were not clearly visible due to the orientation and angle of the organism. The OpenAI o1 has improved performance with a correct identification rate of 50%, although it takes about one minute to output the responses.

Although there may not be much difference in terms of the knowledge each model has about deep-sea organisms, looking at the reasons for the output, o1 tended to include information that was not about the organisms in the image. For example, they identified organisms living in such locations when hydrothermal deposits were present in the background image, and the fact that all the pictures had the notation ©JAMSTEC in the bottom right-hand corner, led to reasons such as 'often seen in JAMSTEC deep-sea images'　and 'as a scene observed by JAMSTEC in the seas around Japan'. were seen. Others chose answers by process of elimination, assuming that no answer was the same as the previous choice, or by associating the answer with the name of the organism. For reference, this benchmark was tested with two human subjects. Although these subjects possessed some knowledge of deep-sea organisms, the percentage of correct answers was approximately 40%. For questions that were unclear, the human subjects often responded by associating the answer with the name of the organism, responding in a similar way to o1 as described above.

On the other hand, the accuracy of the results is likely to be compromised for less well-known species of deep-sea organisms and for species that are challenging to identify due to similarities in the available options.

For example, the output may be depending on the amount of web search hits of *Shinkaia crosnieri* "goemonkoshioriebi" was 4,090, while that of *Kiwa hirsuta*, "yeti crab" was 625,000.

The provision of illustrated books and other materials that are not digital information as external knowledge through RAG (Retrieval Augmented Generation) may also contribute to an improvement in the percentage of correct answers.

### 4 Conclusion

The J-EDI QA was constructed as a benchmark for evaluating the ability to understand multimodal LLM in deep-sea images, particularly in deep-sea organisms. Further details are provided at the end of this paper as a benchmark limited to non-commercial use, including research and education.

An evaluation using Chat GPT with four-choice questions, OpenAI o1, achieved a 50% correct response rate. Moreover, employing questions and answers in Japanese, the identification of deep-sea organisms and comprehension of Japanese terminology were assessed.

The extant models demonstrate a rudimentary comprehension of deep-sea organisms, yet their understanding remains inferior to that of JAMSTEC researchers. To gain a more nuanced understanding of the Japanese language proficiency, it is essential to conduct a parallel assessment utilizing a benchmark translated into English and to analyze the resulting data. To attain greater accuracy, further learning is necessary. It is anticipated that an augmented expert-level dataset will be available in the future and employed for training deep-sea organism-specific LLMs.

As a benchmark, the difficulty level can be increased by increasing the number of choices, as in MMMU-Pro. Additionally, the J-EDI archive contains not only images of deep-sea organisms but also video data, which can be utilized to construct a video dataset in the future.

In addition, it is expected that JAMSTEC will expand beyond deep-sea organisms to include images from other scientific fields in order to establish expert-level benchmarks and develop LLM models tailored to the field of marine and earth sciences.

## Acknowledgements

We express our gratitude to GODAC in JAMSTEC for providing the deep-sea images utilized in the development of the benchmark for this evaluation.

## Data availability

At present, each image data can be obtained from the J-EDI website by agreeing to the terms of use. The 100 image data and QA pairs compiled as J-EDI QA will be made available at a later date, but for now they will be available on individual request.

# Appendix

Table 1: Table of questions, images and answers

| No. | URL | Question and image | Option A | Option B | Option C | Option D | Correct answer |
|---|---|---|---|---|---|---|---|
| 1 | https://www.godac.jamstec.go.jp/jedi/player/j/2K1129IN0099Hp04-22 | 画像の生物は何でしょうか？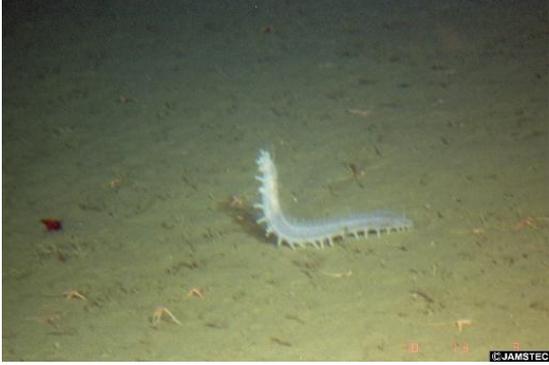 | シンカイハゲナマコ | ウロコムシ | ワタゾコナマコ | センジュナマコ | A |
| 2 | https://www.godac.jamstec.go.jp/jedi/player/j/HPD1014HDTV0126 | 画像の生物は何でしょうか？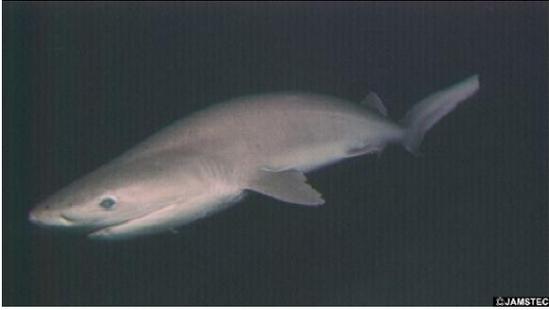 | オンデンザメ | エドアブラザメ | カグラザメ | シロカグラ | C |
| 3 | https://www.godac.jamstec.go.jp/jedi/player/j/2K1395IN0049 | 画像の生物は何でしょうか？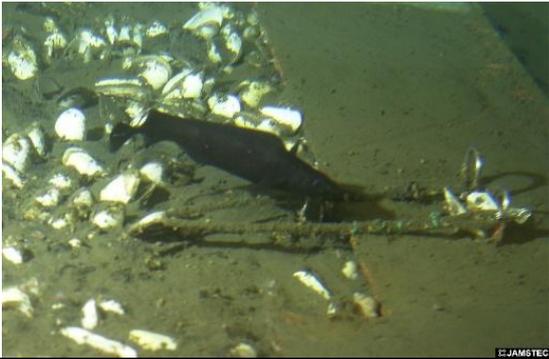 | カナダダラ | カラスダラ | ソコクロダラ | キタノクロダラ | A |
| 4 | https://www.godac.jamstec.go.jp/jedi/player/j/HPD0788HDTV0038 | 画像の生物は何でしょうか？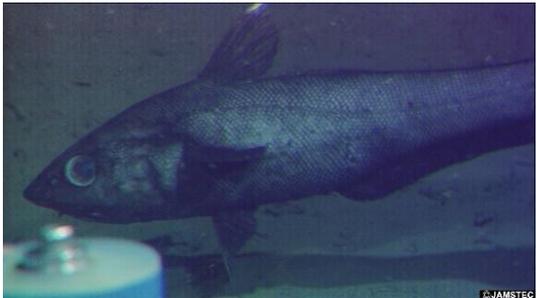 | イバラヒゲ | ヨロイダラ | シンカイヨロイダラ | ムネダラ | A |

Table 1: Table of questions, images and answers (Continue)

| | | 画像の生物は何でしょうか？ | | | | | |
|---|---|---|---|---|---|---|---|
| 5 | https://www.godac.jamstec.go.jp/jedi/player/j/2K0013IN0039Hp02-07 | 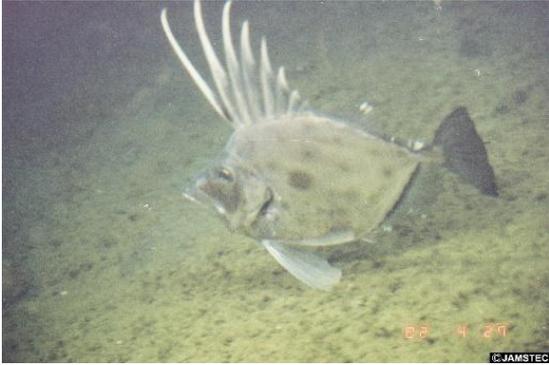 | イトヒキカガミダイ | ミナミカガミダイ | マトウダイ | カガミダイ | D |
| 6 | https://www.godac.jamstec.go.jp/jedi/player/j/2K0730IN0026Hp01-26 | 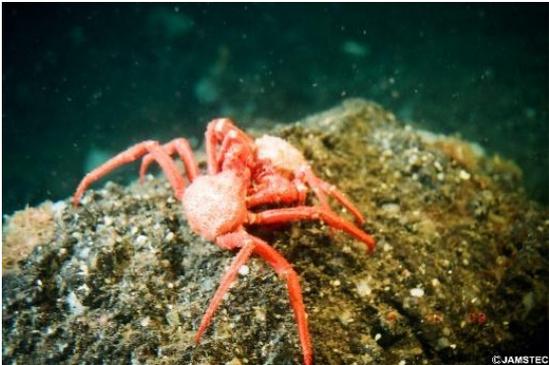 | エゾイバラガニ | ゴカクエゾイバラガニ | ニホンイバラガニ | イバラガニ | A |
| 7 | https://www.godac.jamstec.go.jp/jedi/player/j/HPD1671HDTV1275 | 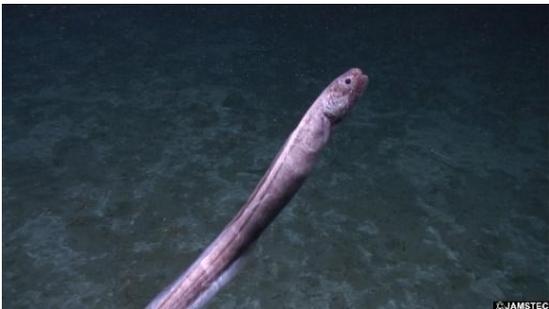 | ホラアナゴ | コンゴウアナゴ | イラコアナゴ | クロアナゴ | B |
| 8 | https://www.godac.jamstec.go.jp/jedi/player/j/2K0709IN0062Hp02-25 | 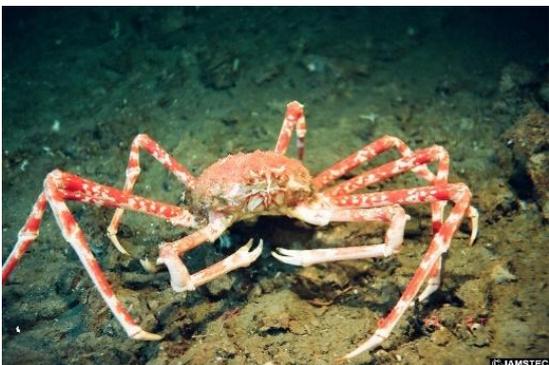 | アシナガツノガニ | オーストンガニ | タカアシガニ | ノコギリガニ | C |

Table 1: Table of questions, images and answers (Continue)

| | | 画像の生物は何でしょうか？ | | | | | |
|---|---|---|---|---|---|---|---|
| 9 | https://www.godac.jamstec.go.jp/jedi/player/j/6K0617IN0038 | 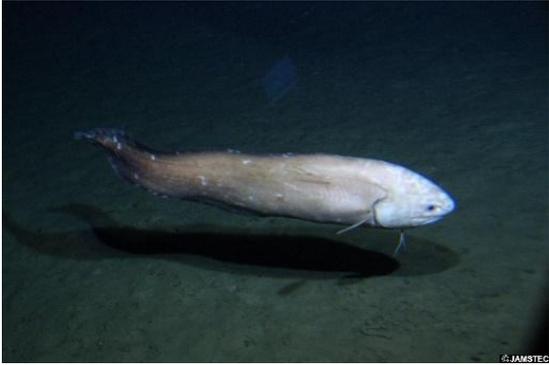 | フクメンイタチウオ | ソコボウズ | ヨミノアシロ | オオイタチウオ | B |
| 10 | https://www.godac.jamstec.go.jp/jedi/player/j/2K0835IN0006Hp01-06 | 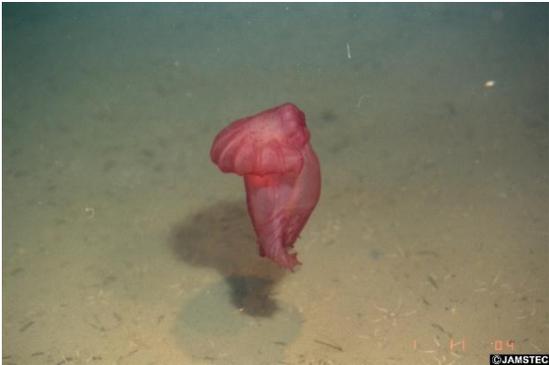 | ハゲナマコ | ウシナマコ | ユメナマコ | クラゲナマコ | C |
| 11 | https://www.godac.jamstec.go.jp/jedi/player/j/HPD1590HDTV0784 | 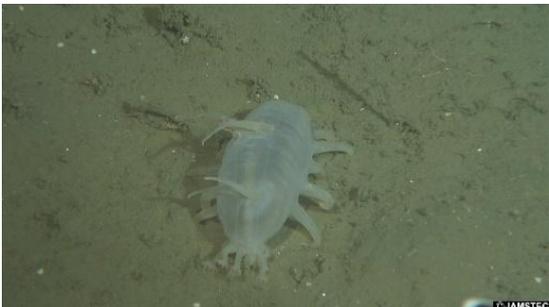 | センジュナマコ | ウシナマコ | ホタテナマコ | ゲジナマコ | A |
| 12 | https://www.godac.jamstec.go.jp/jedi/player/j/HPD0748HDTV1148 | 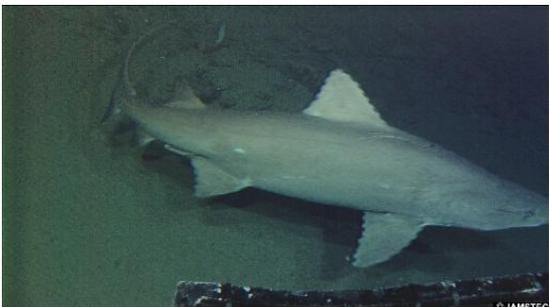 | ミズワニ | シロワニ | ホホジロザメ | オオワニザメ | D |

Table 1: Table of questions, images and answers (Continue)

| | | | | | | | |
|---|---|---|---|---|---|---|---|
| 13 | https://www.godac.jamstec.go.jp/jedi/player/j/HPD0483HDTV0152 | 画像の生物は何でしょうか？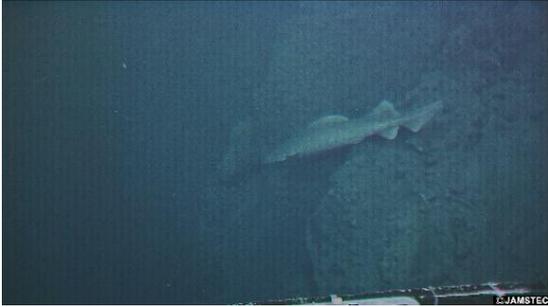 | トガリドチザメ | チヒロザメ | ヒメチヒロザメ | ネムリブカ | B |
| 14 | https://www.godac.jamstec.go.jp/jedi/player/j/6K1293IN0136 | 画像の生物は何でしょうか？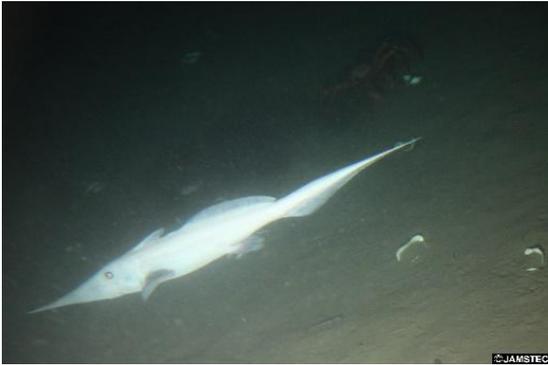 | テングギンザメ | アズマギンザメ | クロテングギンザメ | ヨミノツカイ | A |
| 15 | https://www.godac.jamstec.go.jp/jedi/player/j/2K0730OUT0053Hp03-29 | 画像の生物は何でしょうか？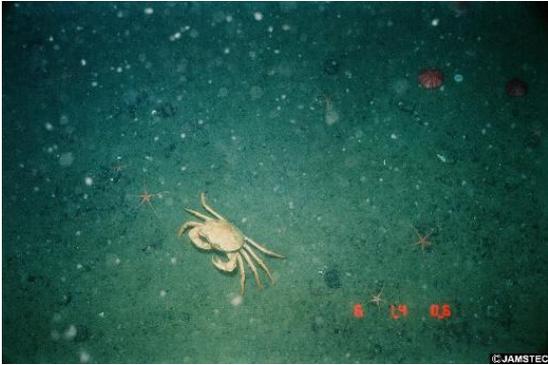 | エンコウガニ | タカアシガニ | オオエンコウガニ | ズワイガニ | C |
| 16 | https://www.godac.jamstec.go.jp/jedi/player/j/2K0550IN0161Hp05-22 | 画像の生物は何でしょうか？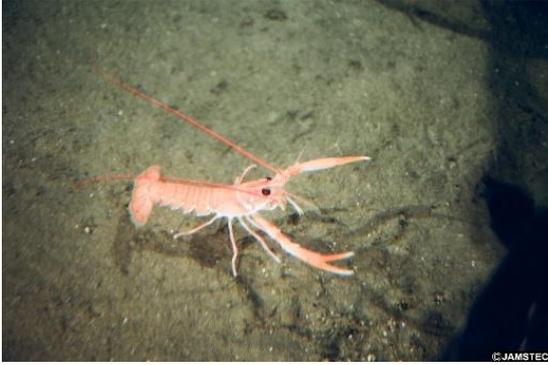 | ミナミアカザエビ | イセエビ | ヒメセンジュエビ | アカザエビ | D |

Table 1: Table of questions, images and answers (Continue)

| | | | | | | | |
|---|---|---|---|---|---|---|---|
| 17 | https://www.godac.jamstec.go.jp/jedi/player/j/2K0808IN0115Hp04-05 | 画像の生物は何でしょうか？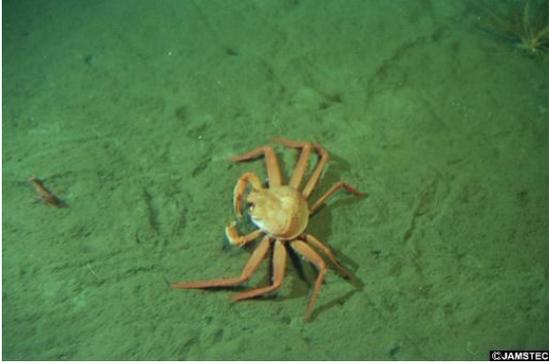 | ベニズワイガニ | ズワイガニ | タカアシガニ | ケガニ | A |
| 18 | https://www.godac.jamstec.go.jp/jedi/player/j/HPD2032HDTV21529 | 画像の生物は何でしょうか？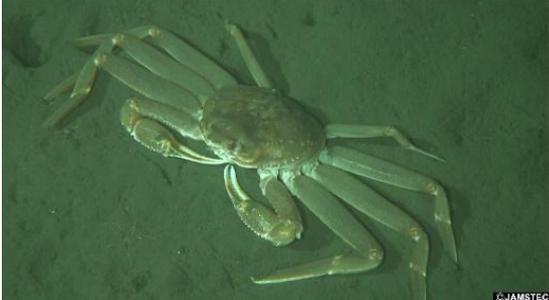 | ベニズワイガニ | ズワイガニ | タカアシガニ | ケガニ | B |
| 19 | https://www.godac.jamstec.go.jp/jedi/player/j/HPD0689OUT0049 | 画像の生物は何でしょうか？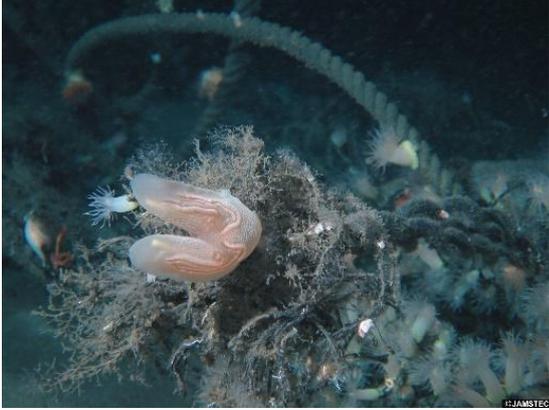 | クラゲムシ | フウセンクラゲ | コトクラゲ | カブトクラゲ | C |
| 20 | https://www.godac.jamstec.go.jp/jedi/player/j/HPD1582HDTV0307 | 画像の生物は何でしょうか？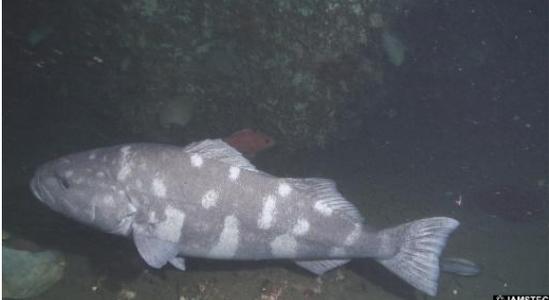 | アブラボウズ | ギンダラ | アイナメ | ホッケ | A |

Table 1: Table of questions, images and answers (Continue)

| | | | | | | | |
|---|---|---|---|---|---|---|---|
| 21 | https://www.godac.jamstec.go.jp/jedi/player/j/2K0025IN0022Hp01-22 | 画像の生物は何でしょうか？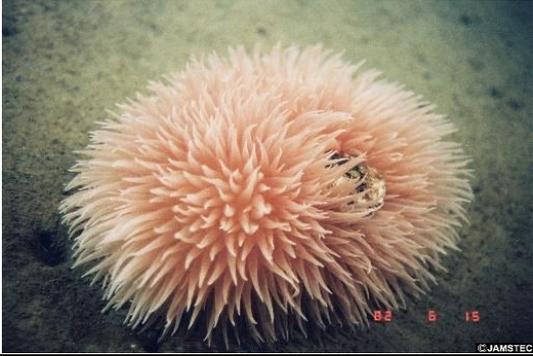 | ダーリアイソギンチャク | ハエジゴクイソギンチャク | セトモノイソギンチャク | ワタゾコイソギンチャク | A |
| 22 | https://www.godac.jamstec.go.jp/jedi/player/j/6K0188IN0034Hp01-34 | 画像の生物は何でしょうか？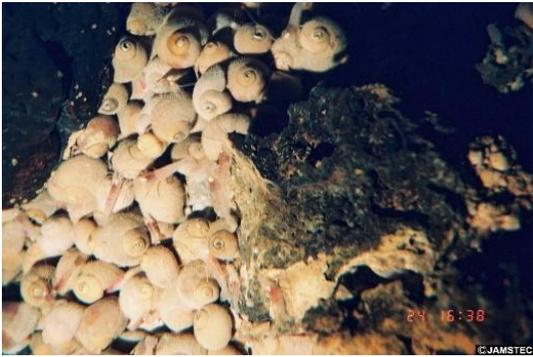 | ヨモツヘグイニナ | ウロコフネタマガイ | サガミハイカブリニナ | アルビンガイ | D |
| 23 | https://www.godac.jamstec.go.jp/jedi/player/j/6K0304OUT0108Hp04-06 | 画像の生物は何でしょうか？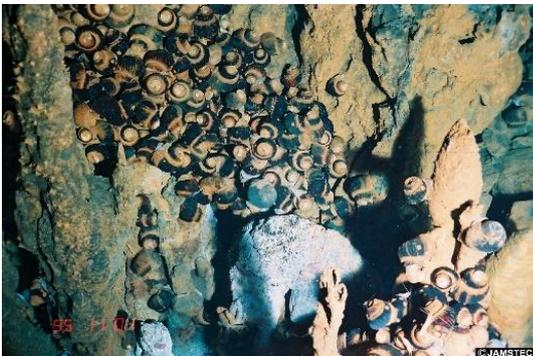 | ヨモツヘグイニナ | ウロコフネタマガイ | サガミハイカブリニナ | アルビンガイ | A |
| 24 | https://www.godac.jamstec.go.jp/jedi/player/j/HPD1677OUT0159 | 画像の生物は何でしょうか？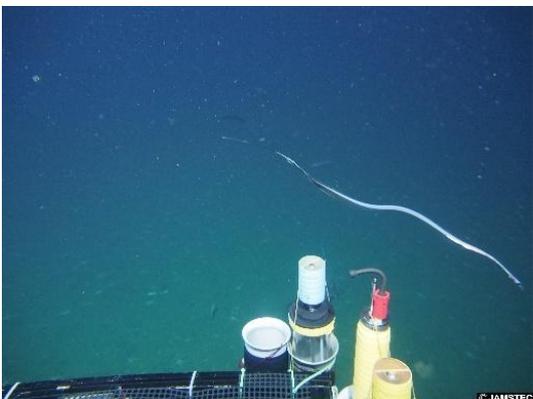 | クロシギウナギ | シギウナギ | クロシギウナギモドキ | ノコバウナギ | B |

Table 1: Table of questions, images and answers (Continue)

| | | | | | | | |
|---|---|---|---|---|---|---|---|
| 25 | https://www.godac.jamstec.go.jp/jedi/player/j/HPD0961OUT0042 | 画像の生物は何でしょうか？<br>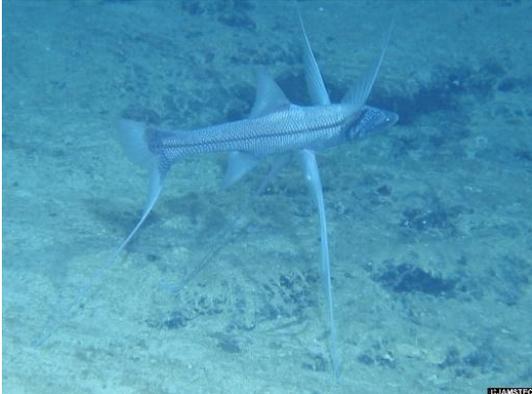 | イトヒキイワシ | チョウチンハダカ | オオイトヒキイワシ | ナガヅエエソ | C |
| 26 | https://www.godac.jamstec.go.jp/jedi/player/j/HPD0821OUT0005 | 画像の生物は何でしょうか？<br>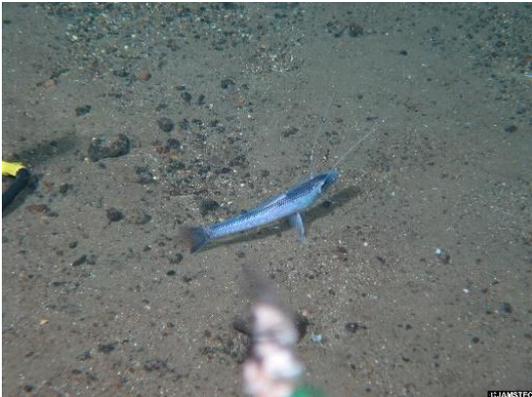 | イトヒキイワシ | チョウチンハダカ | オオイトヒキイワシ | ナガヅエエソ | A |
| 27 | https://www.godac.jamstec.go.jp/jedi/player/j/6K0011IN0035Hp01-35 | 画像の生物は何でしょうか？<br>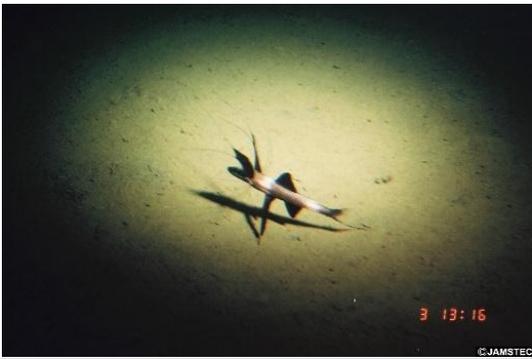 | イトヒキイワシ | チョウチンハダカ | オオイトヒキイワシ | ナガヅエエソ | D |
| 28 | https://www.godac.jamstec.go.jp/jedi/player/j/HPD1537HDTV0556 | 画像の生物は何でしょうか？<br>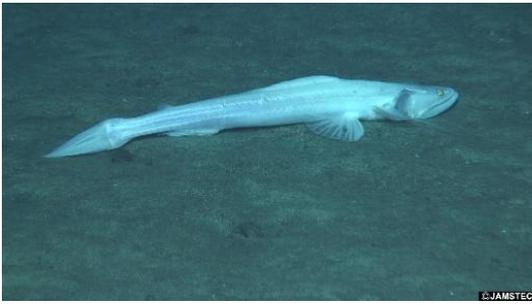 | オニアオメエソ | ミズテング | シンカイエソ | ミズウオ | C |

Table 1: Table of questions, images and answers (Continue)

| | | | | | | | |
|---|---|---|---|---|---|---|---|
| 29 | https://www.godac.jamstec.go.jp/jedi/player/j/HPD1586HDTV0097 | 画像の生物は何でしょうか？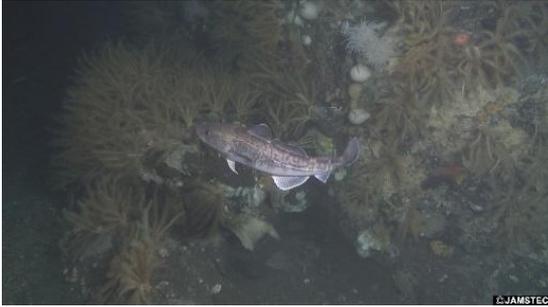 | マダラ | スケトウダラ | チゴダラ | カナダダラ | A |
| 30 | https://www.godac.jamstec.go.jp/jedi/player/j/2K0711IN0084Hp03-10 | 画像の生物は何でしょうか？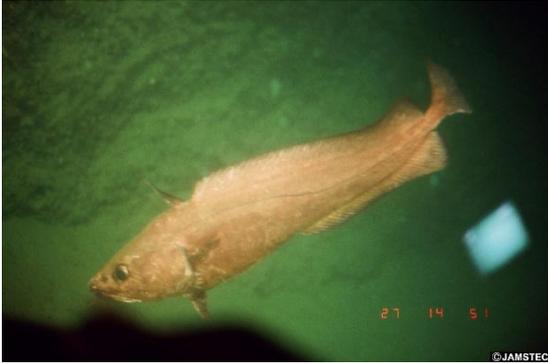 | マダラ | スケトウダラ | チゴダラ | カナダダラ | C |
| 31 | https://www.godac.jamstec.go.jp/jedi/player/j/HPD1668HDTV2210 | 画像の生物は何でしょうか？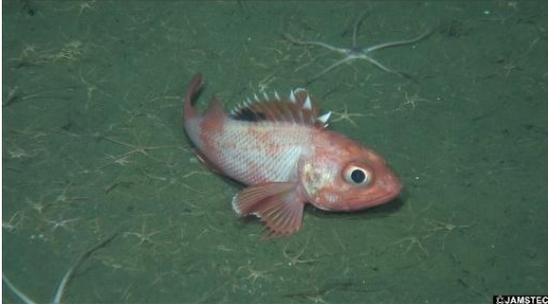 | アコウダイ | ホウズキ | キチジ | アラメヌケ | C |
| 32 | https://www.godac.jamstec.go.jp/jedi/player/j/2K1290IN0020 | 画像の生物は何でしょうか？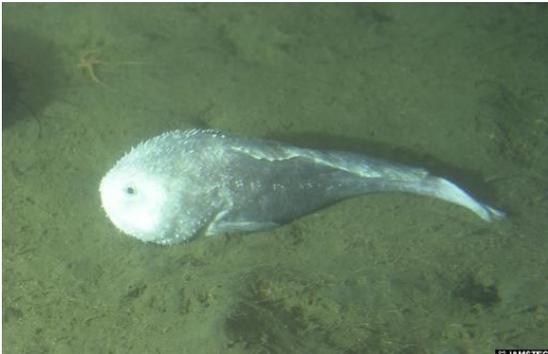 | コブシカジカ | コンニャクカジカ | ボウズカジカ | ニュウドウカジカ | D |

Table 1: Table of questions, images and answers (Continue)

| | | 画像の生物は何でしょうか？ | | | | | |
|---|---|---|---|---|---|---|---|
| 33 | https://www.godac.jamstec.go.jp/jedi/player/j/2K0124IN0036Hp02-12 | 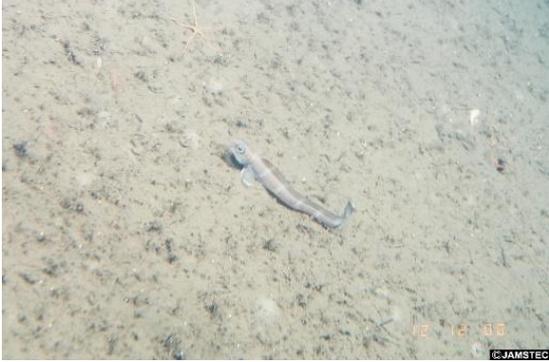 | タナカゲンゲ | カンテンゲンゲ | ヨコスジクロゲンゲ | ノロゲンゲ | C |
| 34 | https://www.godac.jamstec.go.jp/jedi/player/j/HPD0527HDTV0120 | 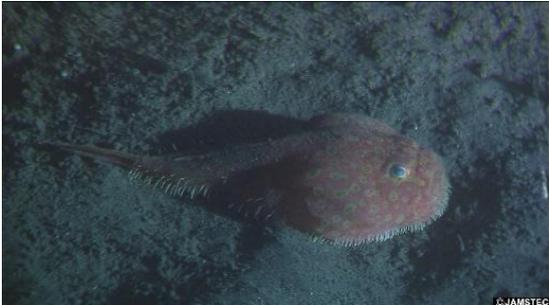 | ミドリフサアンコウ | アカグツ | キアンコウ | ハナグロフサアンコウ | A |
| 35 | https://www.godac.jamstec.go.jp/jedi/player/j/6K1033IN0188 | 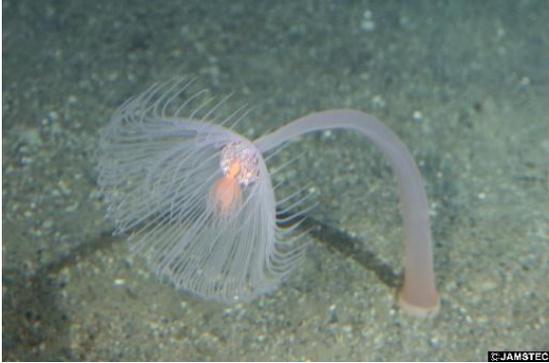 | カタアシクラゲ | オトヒメノハナガサ | フカウラヒドラ | トモオオウミヒドラ | B |
| 36 | https://www.godac.jamstec.go.jp/jedi/player/j/2K1409IN0132 | 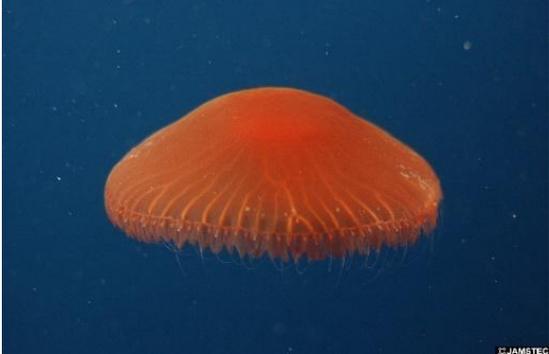 | ディープスタリアクラゲ | ユビアシクラゲ | リンゴクラゲ | ダイオウクラゲ | C |

Table 1: Table of questions, images and answers (Continue)

| | | | | | | | |
|---|---|---|---|---|---|---|---|
| 37 | https://www.godac.jamstec.go.jp/jedi/player/j/HPD0352HDTV0549 | 画像の生物は何でしょうか？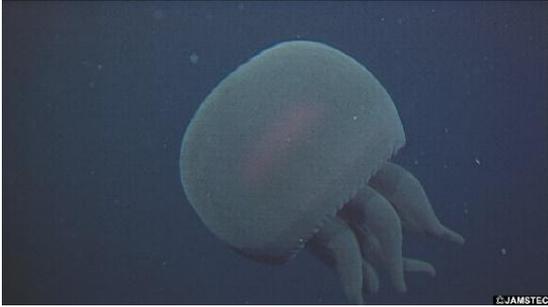 | ディープスタリアクラゲ | ユビアシクラゲ | リンゴクラゲ | ダイオウクラゲ | B |
| 38 | https://www.godac.jamstec.go.jp/jedi/player/j/HPD0313HDTV0153 | 画像の生物は何でしょうか？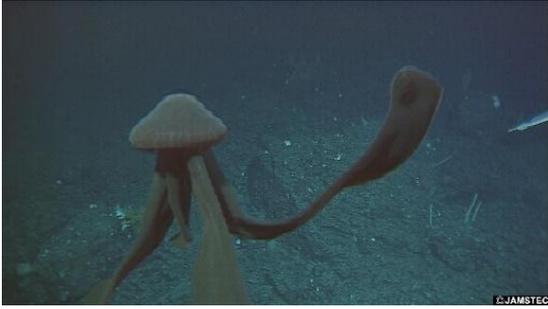 | ディープスタリアクラゲ | ユビアシクラゲ | リンゴクラゲ | ダイオウクラゲ | D |
| 39 | https://www.godac.jamstec.go.jp/jedi/player/j/HPD0613HDTV0055 | 画像の生物は何でしょうか？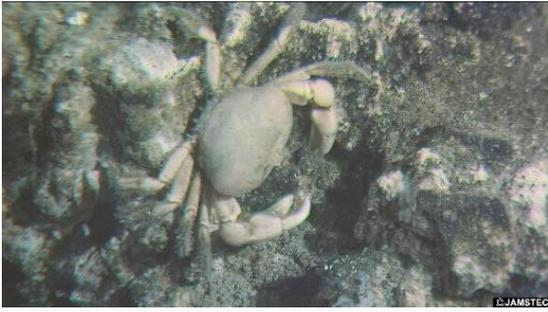 | ユノハナガニ | ゴエモンコシオリエビ | オオエンコウガニ | オウギガニ | A |
| 40 | https://www.godac.jamstec.go.jp/jedi/player/j/HPD0399OUT0130 | 画像の生物は何でしょうか？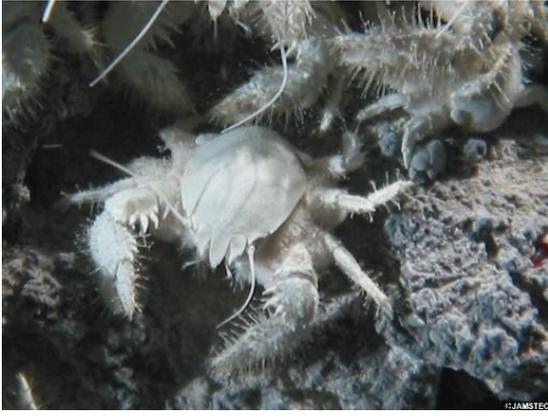 | シンカイコシオリエビ | ゴエモンコシオリエビ | オオコシオリエビ | イエティクラブ | B |

Table 1: Table of questions, images and answers (Continue)

| | | | | | | | |
|---|---|---|---|---|---|---|---|
| 41 | https://www.godac.jamstec.go.jp/jedi/player/j/2K0583IN0072Hp03-04 | 画像の生物は何でしょうか？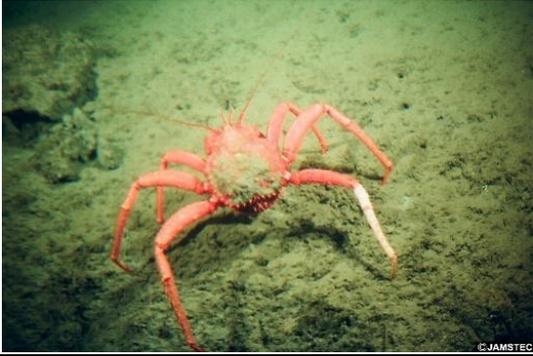 | エゾイバラガニ | タカアシガニ | ニホンイバラガニ | タラバガニ | C |
| 42 | https://www.godac.jamstec.go.jp/jedi/player/j/HPD0399OUT0013 | 画像の生物は何でしょうか？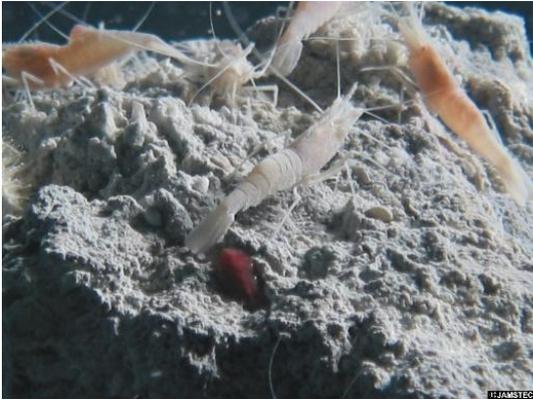 | ゴエモンコシオリエビ | ジゴクモエビ | カイレイツノナシオハラエビ | オハラエビ | D |
| 43 | https://www.godac.jamstec.go.jp/jedi/player/j/2K0580IN0072Hp02-36 | 画像の生物は何でしょうか？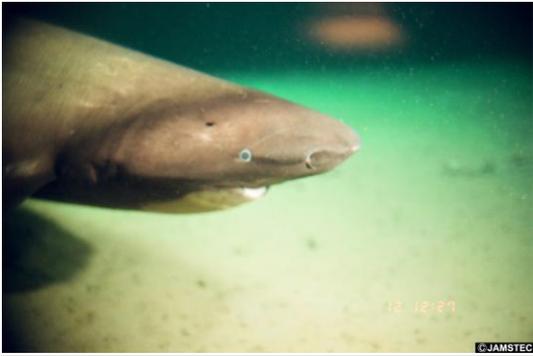 | オンデンザメ | エドアブラザメ | カグラザメ | シロカグラ | A |
| 44 | https://www.godac.jamstec.go.jp/jedi/player/j/6K0129OUT0020Hp01-20 | 画像の生物は何でしょうか？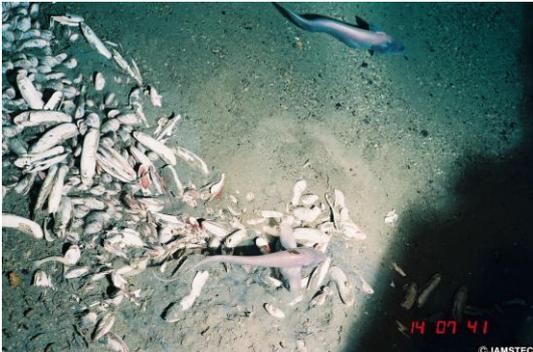 | イバラヒゲ | ヨロイダラ | シンカイヨロイダラ | ムネダラ | C |

Table 1: Table of questions, images and answers (Continue)

| 45 | https://www.godac.jamstec.go.jp/jedi/player/j/2K1184IN0075Hp03-08 | 画像の生物は何でしょうか？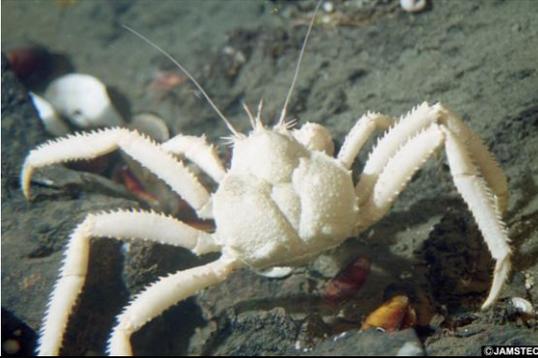 | エゾイバラガニ | ゴカクエゾイバラガニ | ニホンイバラガニ | イバラガニ | B |
| --- | --- | --- | --- | --- | --- | --- | --- |
| 46 | https://www.godac.jamstec.go.jp/jedi/player/j/2K0130IN0005Hp01-05 | 画像の生物は何でしょうか？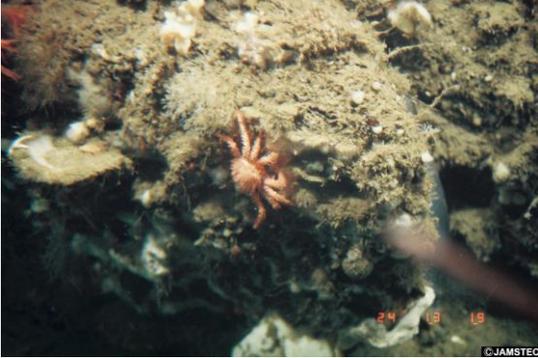 | ハリイバラガニ | イバラガニモドキ | ニホンイバラガニ | イバラガニ | B |
| 47 | https://www.godac.jamstec.go.jp/jedi/player/j/2K1058IN0002Hp01-03 | 画像の生物は何でしょうか？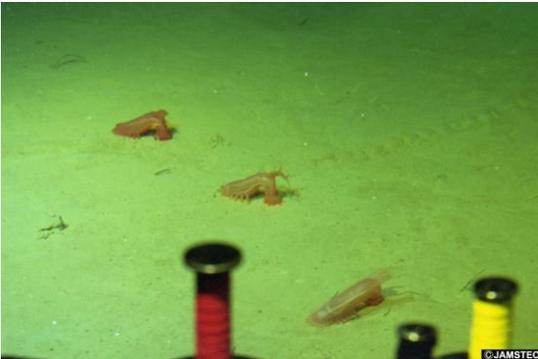 | ハゲナマコ | キャラウシナマコ | ウカレウシナマコ | クラゲナマコ | C |
| 48 | https://www.godac.jamstec.go.jp/jedi/player/j/6K0614IN0005 | 画像の生物は何でしょうか？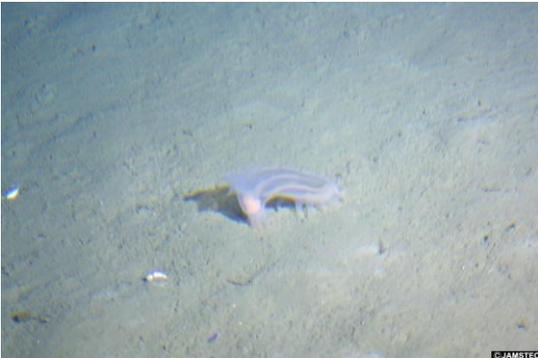 | ハゲナマコ | キャラウシナマコ | ウカレウシナマコ | クラゲナマコ | B |

Table 1: Table of questions, images and answers (Continue)

| | | 画像の生物は何でしょうか？ | | | | | | |
|---|---|---|---|---|---|---|---|---|
| 49 | https://www.godac.jamstec.go.jp/jedi/player/j/6K0190OUT0145Hp05-01 | 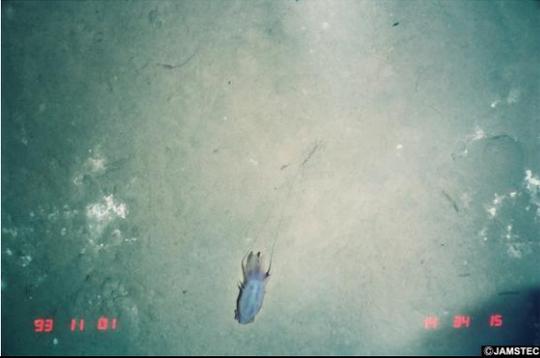 | | センジュナマコ | ウシナマコ | ホタテナマコ | ゲジナマコ | C |
| 50 | https://www.godac.jamstec.go.jp/jedi/player/j/6K0674IN0053 | 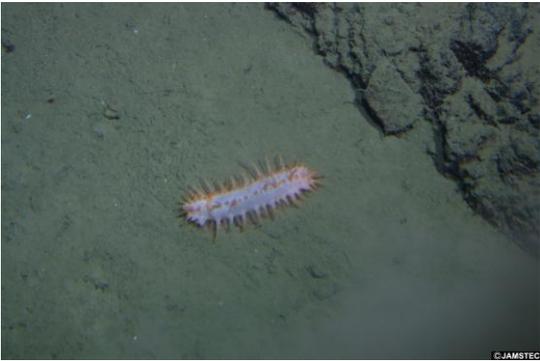 | | センジュナマコ | ウシナマコ | ホタテナマコ | ゲジナマコ | D |
| 51 | https://www.godac.jamstec.go.jp/jedi/player/j/2K1338IN0194 | 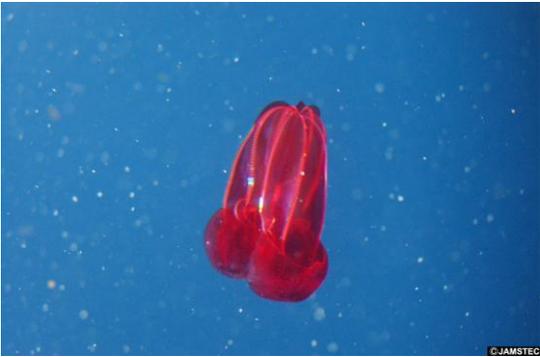 | | クラゲムシ | フウセンクラゲ | コトクラゲ | アカカブトクラゲ | D |
| 52 | https://www.godac.jamstec.go.jp/jedi/player/j/2K1278IN0039 | 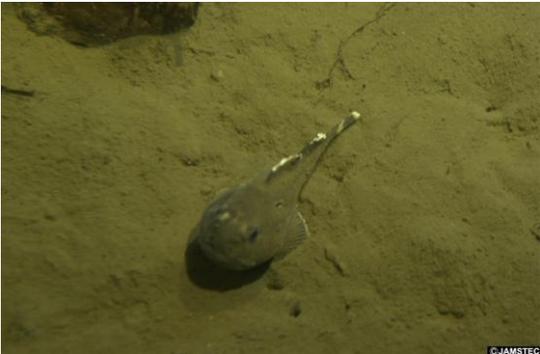 | | ヤマトコブシカジカ | コンニャクカジカ | ボウズカジカ | ニュウドウカジカ | A |

Table 1: Table of questions, images and answers (Continue)

| | | | | | | | |
|---|---|---|---|---|---|---|---|
| 53 | https://www.godac.jamstec.go.jp/jedi/player/j/HPD0434HDTV0784 | 画像の生物は何でしょうか？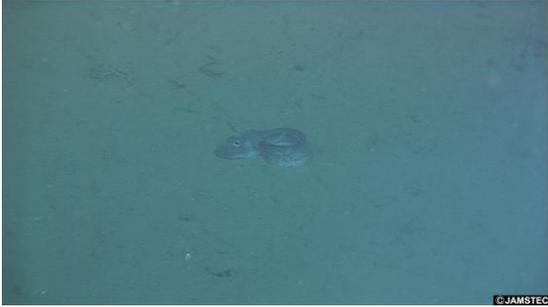 | タナカゲンゲ | カンテンゲンゲ | ヨコスジクロゲンゲ | ノロゲンゲ | A |
| 54 | https://www.godac.jamstec.go.jp/jedi/player/j/2K1280IN0048 | 画像の生物は何でしょうか？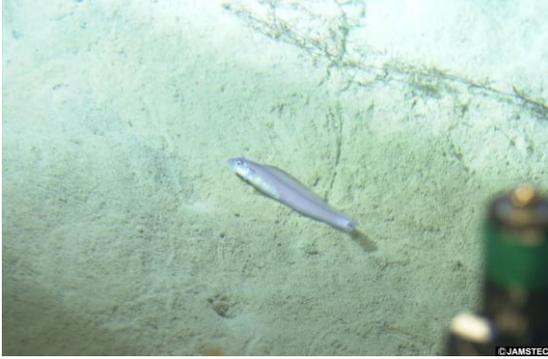 | タナカゲンゲ | カンテンゲンゲ | ヨコスジクロゲンゲ | ノロゲンゲ | D |
| 55 | https://www.godac.jamstec.go.jp/jedi/player/j/3K0426ASC0072Hp02-36 | 画像の生物は何でしょうか？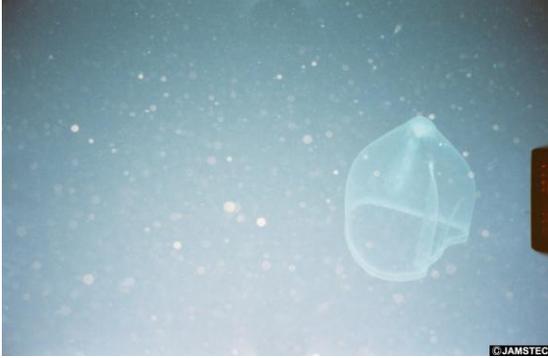 | ディープスタリアクラゲ | ユビアシクラゲ | リンゴクラゲ | ダイオウクラゲ | A |
| 56 | https://www.godac.jamstec.go.jp/jedi/player/j/2K1275IN0030 | 画像の生物は何でしょうか？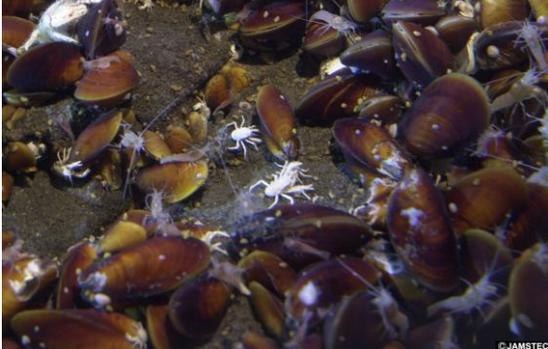 | ジゴクモエビ | リュウキュウシンカイコシオリエビ | オオコシオリエビ | イエティクラブ | B |

Table 1: Table of questions, images and answers (Continue)

| | | | | | | | |
|---|---|---|---|---|---|---|---|
| 57 | https://www.godac.jamstec.go.jp/jedi/player/j/2K0470IN0032Hp01-32 | 画像の生物は何でしょうか？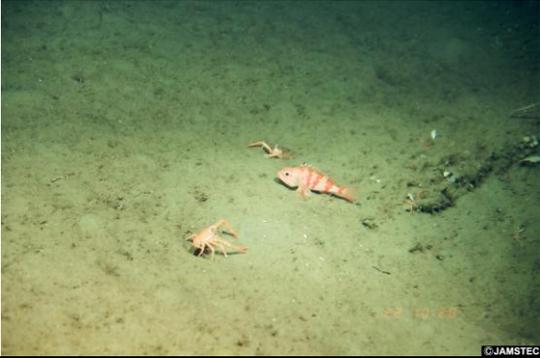 | ジゴクモエビ | リュウキュウシンカイコシオリエビ | オオコシオリエビ | イエティクラブ | C |
| 58 | https://www.godac.jamstec.go.jp/jedi/player/j/6K0660IN0066 | 画像の生物は何でしょうか？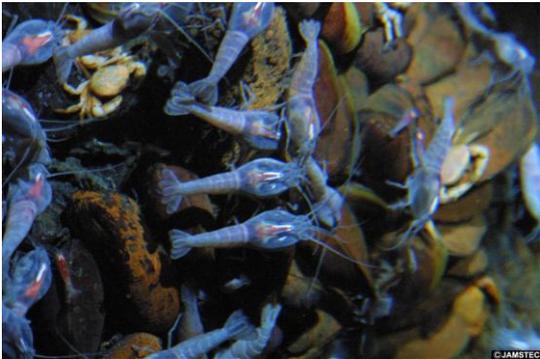 | ゴエモンコシオリエビ | ジゴクモエビ | カイレイツノナシオハラエビ | オハラエビ | C |
| 59 | https://www.godac.jamstec.go.jp/jedi/player/j/HPD0700OUT0041 | 画像の生物は何でしょうか？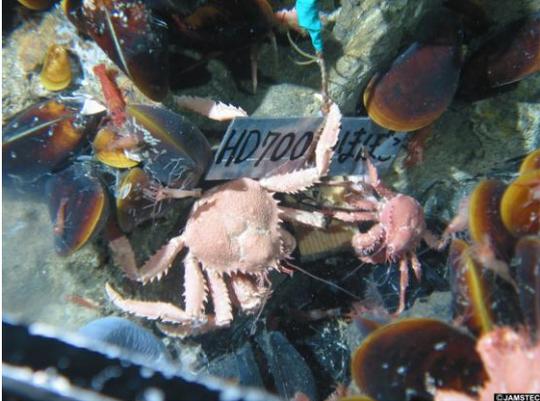 | エンセイエゾイバラガニ | タカアシガニ | ニホンイバラガニ | タラバガニ | A |
| 60 | https://www.godac.jamstec.go.jp/jedi/player/j/6K1250IN0036 | 画像の生物は何でしょうか？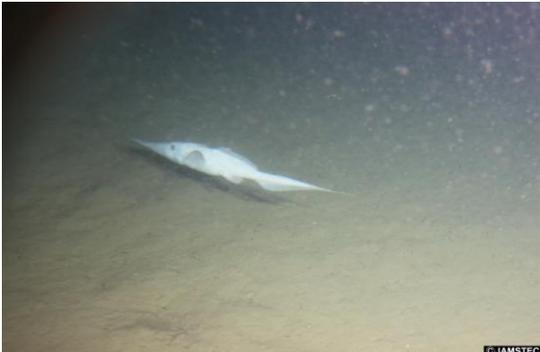 | テングギンザメ科 | アズマギンザメ | ギンザメ | ココノホシギンザメ | A |

Table 1: Table of questions, images and answers (Continue)

| | | 画像の生物は何でしょうか？ | | | | | |
|---|---|---|---|---|---|---|---|
| 61 | https://www.godac.jamstec.go.jp/jedi/player/j/HPD0352HDTV0694 | 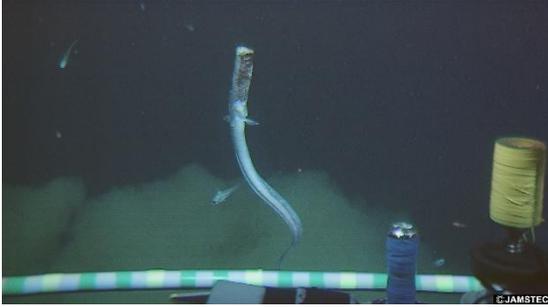 | コンゴウアナゴ | ヌタウナギ科 | ホラアナゴ科 | イラコアナゴ | C |
| 62 | https://www.godac.jamstec.go.jp/jedi/player/j/2K0136IN0013Hp01-13 | 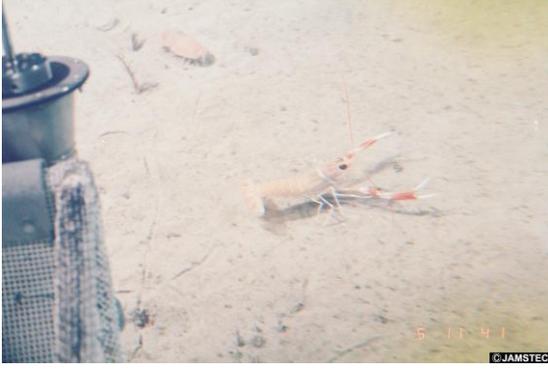 | サガミアカザエビ | ホソウデヤスリアカザエビ | ハサミモエビ | イバラモエビ | A |
| 63 | https://www.godac.jamstec.go.jp/jedi/player/j/HPD1106HDTV0083 | 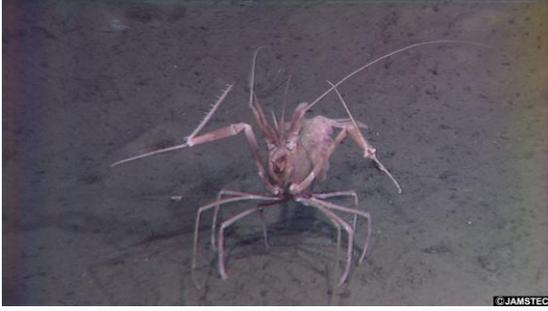 | サガミアカザエビ | ホソウデヤスリアカザエビ | ハサミモエビ | イバラモエビ | B |
| 64 | https://www.godac.jamstec.go.jp/jedi/player/j/2K0637IN0107Hp03-33 | 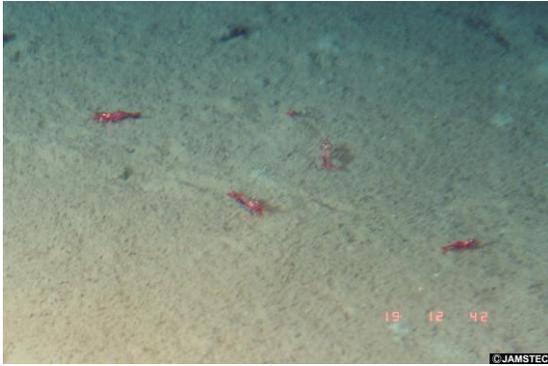 | サガミアカザエビ | ホソウデヤスリアカザエビ | ハサミモエビ | イバラモエビ | C |

Table 1: Table of questions, images and answers (Continue)

| | | | | | | | | |
|---|---|---|---|---|---|---|---|---|
| 65 | https://www.godac.jamstec.go.jp/jedi/player/j/2K0506IN0220Hp07-10 | 画像の生物は何でしょうか？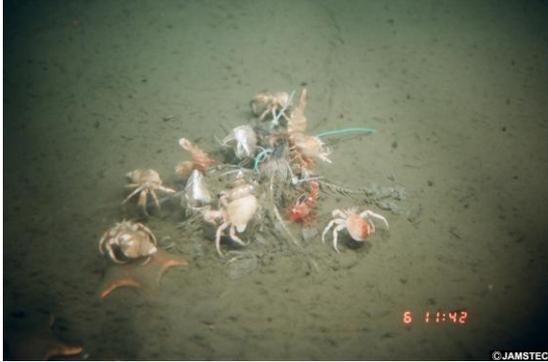 | サガミアカザエビ | ホソウデヤスリアカザエビ | ハサミモエビ | イバラモエビ | D |
| 66 | https://www.godac.jamstec.go.jp/jedi/player/j/HPD0351HDTV0781 | 画像の生物は何でしょうか？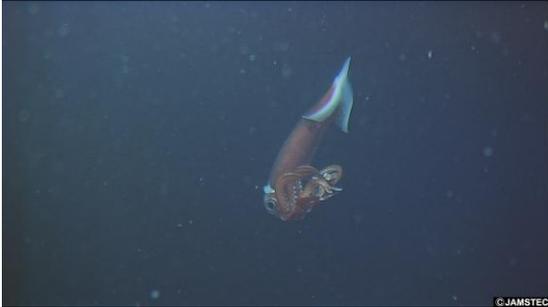 | ユウレイイカ | マッコウタコイカ | テナガタコイカ | テカギイカ | D |
| 67 | https://www.godac.jamstec.go.jp/jedi/player/j/2K0186IN0004Hp01-04 | 画像の生物は何でしょうか？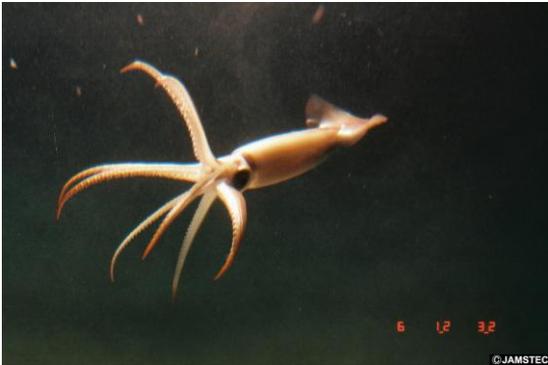 | ユウレイイカ | マッコウタコイカ | テナガタコイカ | テカギイカ | C |
| 68 | https://www.godac.jamstec.go.jp/jedi/player/j/2K0742IN0159Hp06-10 | 画像の生物は何でしょうか？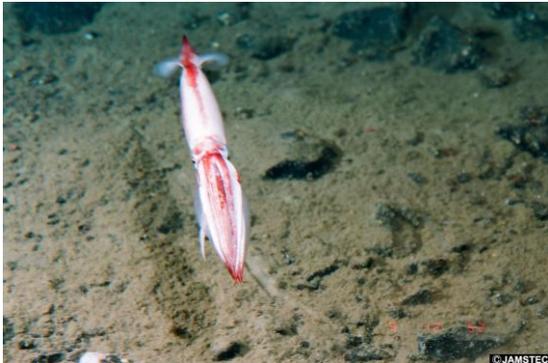 | ユウレイイカ | マッコウタコイカ | テナガタコイカ | テカギイカ | B |

Table 1: Table of questions, images and answers (Continue)

| 69 | https://www.godac.jamstec.go.jp/jedi/player/j/2K0101IN0106Hp05-13 | 画像の生物は何でしょうか？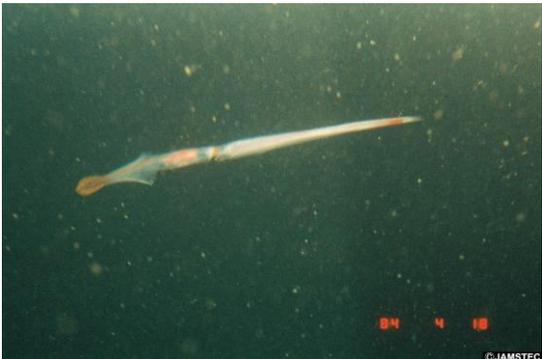 | ユウレイイカ | マッコウタコイカ | テナガタコイカ | テカギイカ | A |
| 70 | https://www.godac.jamstec.go.jp/jedi/player/j/6K0690IN0052 | 画像の生物は何でしょうか？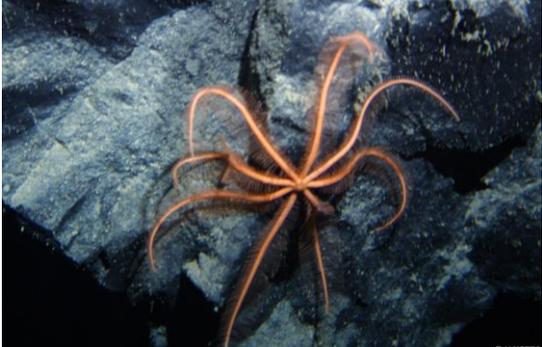 | クモヒトデ | ハネウデボソヒトデ科 | ゴカクヒトデ | シワウデボソヒトデ科 | D |
| 71 | https://www.godac.jamstec.go.jp/jedi/player/j/6K0359OUT0061Hp02-25 | 画像の生物は何でしょうか？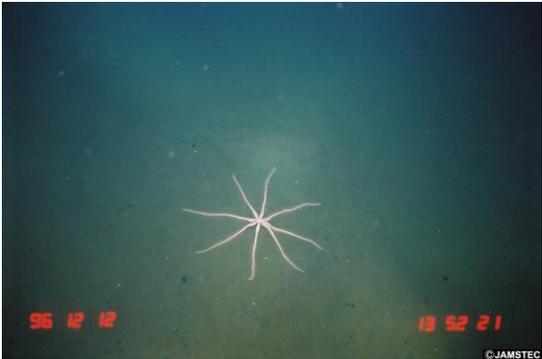 | クモヒトデ | ハネウデボソヒトデ科 | ゴカクヒトデ | シワウデボソヒトデ科 | B |
| 72 | https://www.godac.jamstec.go.jp/jedi/player/j/2K0260OUT0024Hp01-24 | 画像の生物は何でしょうか？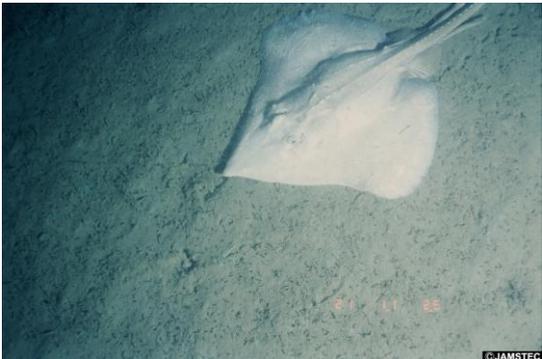 | ムツエラエイ | ウスエイ | チヒロカスベ | アリューシャンカスベ | C |

Table 1: Table of questions, images and answers (Continue)

| | | | | | | | |
|---|---|---|---|---|---|---|---|
| 73 | https://www.godac.jamstec.go.jp/jedi/player/j/6K0353OUT0155Hp08-11 | 画像の生物は何でしょうか？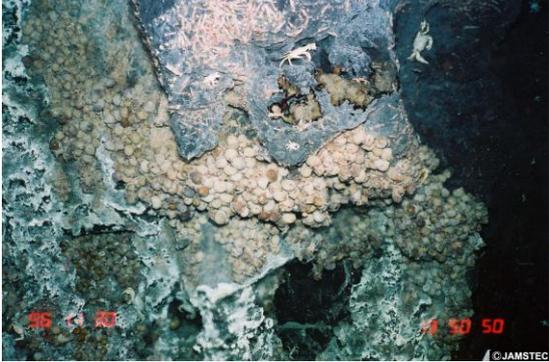 | ヘイトウシンカイヒバリガイ | アルビンガイ | ヨモツヘグイニナ | ナギナタシロウリガイ | B |
| 74 | https://www.godac.jamstec.go.jp/jedi/player/j/6K0144OUT0176Hp06-32 | 画像の生物は何でしょうか？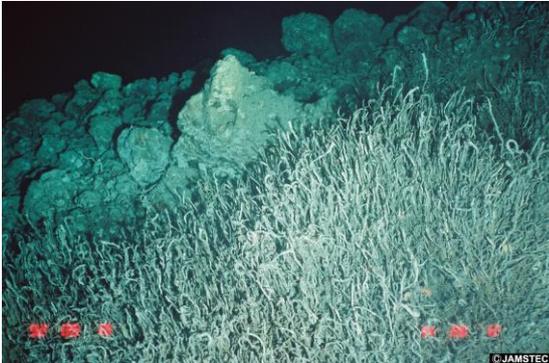 | ホネクイハナムシ | ガラパゴスハオリムシ | サツマハオリムシ | エジソンレパス | C |
| 75 | https://www.godac.jamstec.go.jp/jedi/player/j/6K0383IN0036Hp01-36 | 画像の生物は何でしょうか？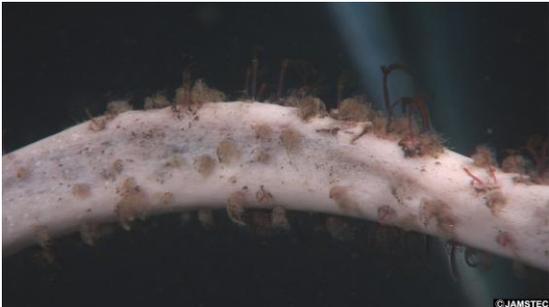 | ホネクイハナムシ | ガラパゴスハオリムシ | サツマハオリムシ | エジソンレパス | B |
| 76 | https://www.godac.jamstec.go.jp/jedi/player/j/HPD0300HDTV0207 | 画像の生物は何でしょうか？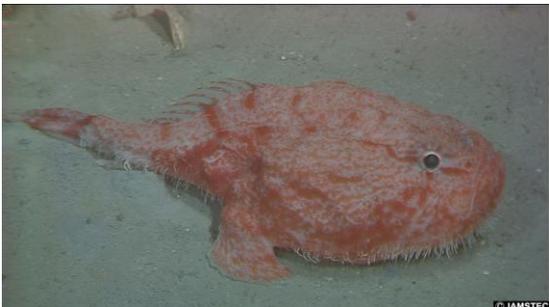 | ホンフサアンコウ | ミドリフサアンコウ | キアンコウ | ハナグロフサアンコウ | D |

Table 1: Table of questions, images and answers (Continue)

| | | | | | | | | |
|---|---|---|---|---|---|---|---|---|
| 77 | https://www.godac.jamstec.go.jp/jedi/player/j/HPD1382OUT0016 | 画像の生物は何でしょうか？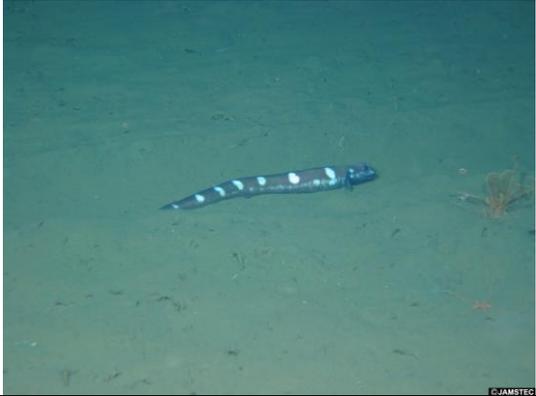 | シロブチヘビゲンゲ | ヨコスジクロゲンゲ | リュウキュウヘビゲンゲ | ダイダイヘビゲンゲ | A |
| 78 | https://www.godac.jamstec.go.jp/jedi/player/j/HPD1412HDTV0056 | 画像の生物は何でしょうか？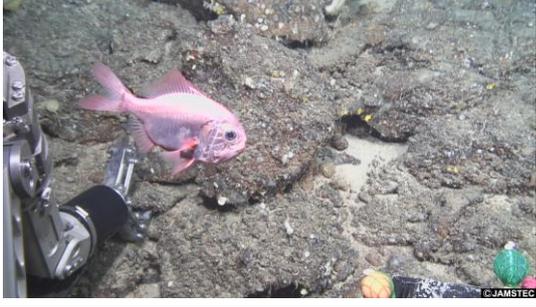 | クロヒウチ | マルヒウチダイ | ハシキンメ | ハリダシエビス | B |
| 79 | https://www.godac.jamstec.go.jp/jedi/player/j/HPD1118OUT0137 | 画像の生物は何でしょうか？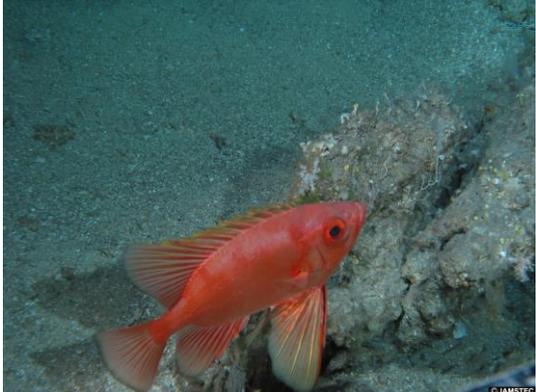 | ゴマヒレキントキ | クルマダイ | ミナミキントキ | チカメキントキ | D |
| 80 | https://www.godac.jamstec.go.jp/jedi/player/j/2K0133IN0053Hp02-16 | 画像の生物は何でしょうか？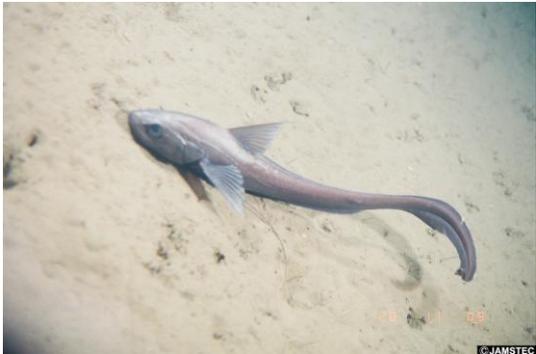 | キタノソコダラ | ヘリダラ | ヒモダラ | ハナソコダラ | C |

Table 1: Table of questions, images and answers (Continue)

| | | 画像の生物は何でしょうか？ | | | | | |
|---|---|---|---|---|---|---|---|
| 81 | https://www.godac.jamstec.go.jp/jedi/player/j/2K0124IN0036Hp02-12 | 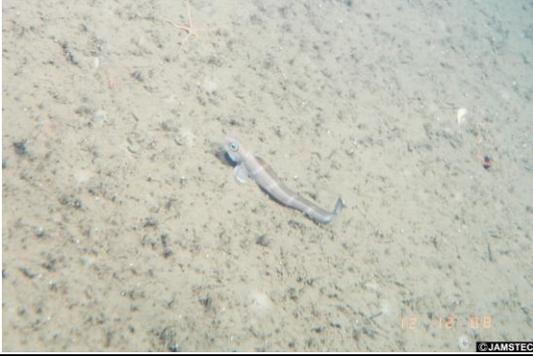 | シロブチヘビゲンゲ | ヨコスジクロゲンゲ | リュウキュウヘビゲンゲ | ダイダイヘビゲンゲ | B |
| 82 | https://www.godac.jamstec.go.jp/jedi/player/j/HPD0448OUT0013 | 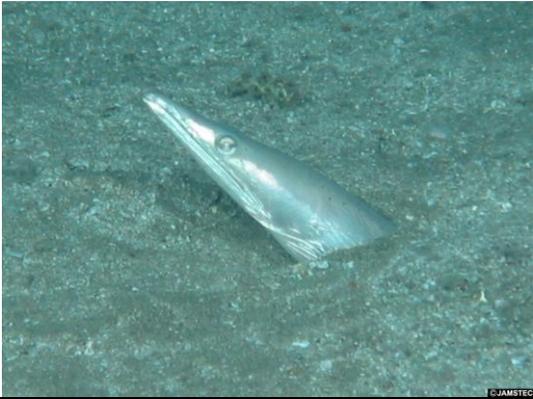 | ムカシウミヘビ | ミミズアナゴ | セレベスヒゲアナゴ | ダイナンウミヘビ | D |
| 83 | https://www.godac.jamstec.go.jp/jedi/player/j/2K0981OUT0047Hp03-17 | 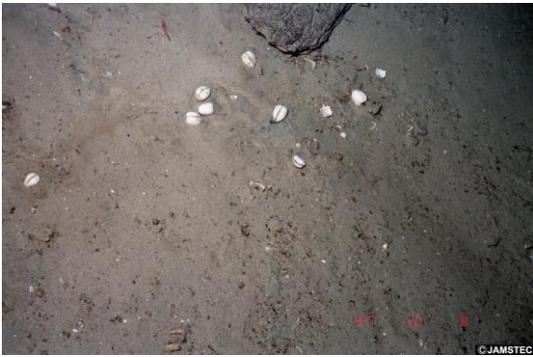 | カイコウオトヒメハマグリ | ナカイオトヒメハマグリ | オトヒメハマグリ | クロシマオトヒメハマグリ | D |
| 84 | https://www.godac.jamstec.go.jp/jedi/player/j/HPD0623OUT0120 | 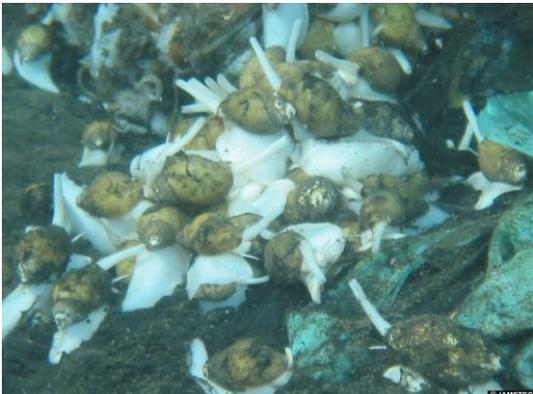 | ヘイトウシンカイヒバリガイ | シチヨウシンカイヒバリガイ | ソウヨウバイ | ツブナリシャジク | D |

Table 1: Table of questions, images and answers (Continue)

| | | | | | | | | |
|---|---|---|---|---|---|---|---|---|
| 85 | https://www.godac.jamstec.go.jp/jedi/player/j/2K1327IN0027 | 画像の生物は何でしょうか？<br />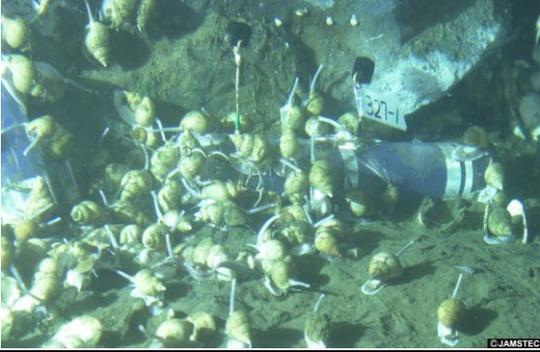 | ヘイトウシンカイヒバリガイ | シチョウシンカイヒバリガイ | ソウヨウバイ | ツブナリシャジク | C |
| 86 | https://www.godac.jamstec.go.jp/jedi/player/j/6K0557IN0009Hp01-09 | 画像の生物は何でしょうか？<br />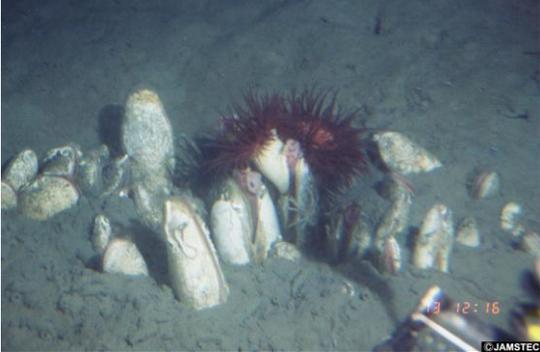 | ナギナタシロウリガイ | ツバサシロウリガイ | ヘイトウシンカイヒバリガイ | シチョウシンカイヒバリガイ | B |
| 87 | https://www.godac.jamstec.go.jp/jedi/player/j/6K0682IN0063 | 画像の生物は何でしょうか？<br />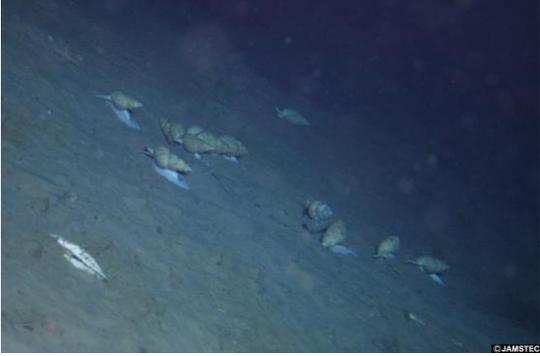 | キノシタバイ | カイコウツムバイ | スエヒロキヌタレガイ | ダマシエヒザラガイ | A |
| 88 | https://www.godac.jamstec.go.jp/jedi/player/j/6K0672IN0015 | 画像の生物は何でしょうか？<br />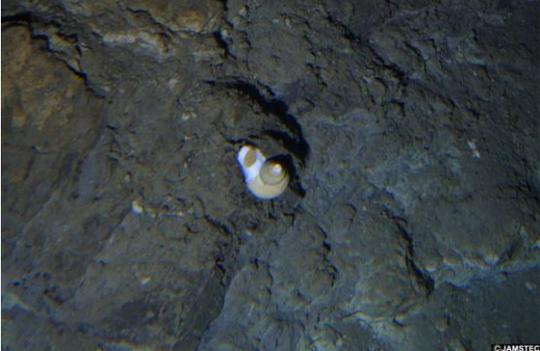 | キノシタバイ | カイコウツムバイ | スエヒロキヌタレガイ | ダマシエヒザラガイ | B |

Table 1: Table of questions, images and answers (Continue)

| 89 | https://www.godac.jamstec.go.jp/jedi/player/j/HPD0188HDTV0283 | 画像の生物は何でしょうか？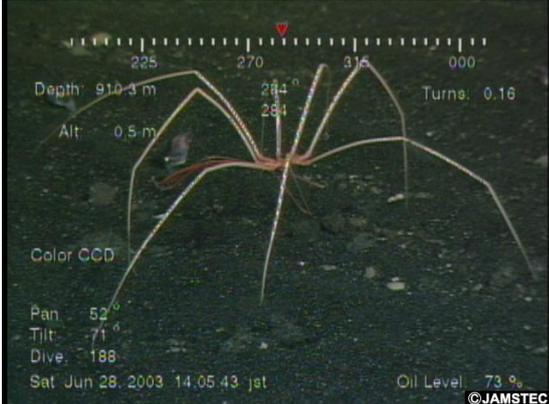 | ナスタオオウミグモ | ベニオオウミグモ | チビオクダウミグモ | チビカニノテウミグモ | A |
|---|---|---|---|---|---|---|---|
| 90 | https://www.godac.jamstec.go.jp/jedi/player/j/2K1283IN0060 | 画像の生物は何でしょうか？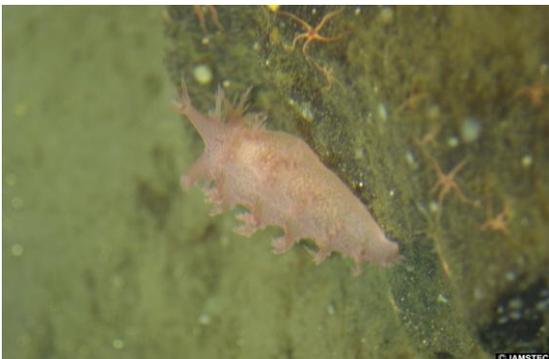 | ミドリウミウシ属 | キイロハナガサウミウシ属 | ホクヨウミウシ属 | ユビノウハナガサウミウシ | C |
| 91 | https://www.godac.jamstec.go.jp/jedi/player/j/2K1286IN0010 | 画像の生物は何でしょうか？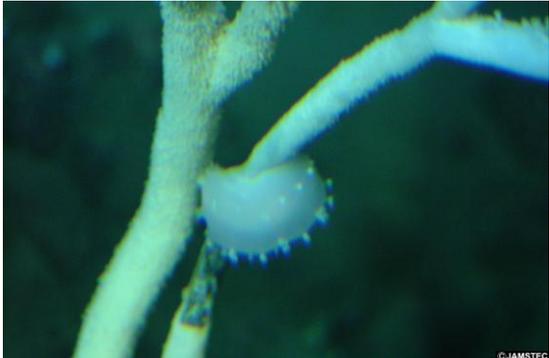 | シロホクヨウミウシ | シラヒメハナガサウミウシ | キイロハナガサウミウシ | ユビノウハナガサウミウシ | A |
| 92 | https://www.godac.jamstec.go.jp/jedi/player/j/HPD0439HDTV0095 | 画像の生物は何でしょうか？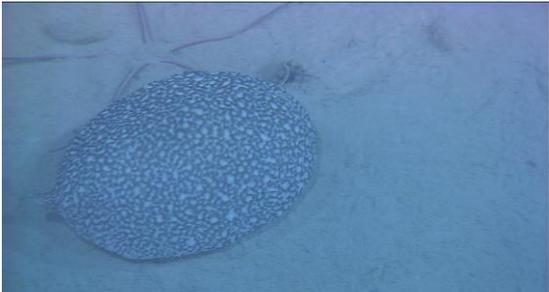 | サラサベッコウタマガイ | イボベッコウタマガイ | セトベッコウタマガイ | キシュウベッコウタマガイ | A |

Table 1: Table of questions, images and answers (Continue)

| | | 画像の生物は何でしょうか？ | | | | | |
|---|---|---|---|---|---|---|---|
| 93 | https://www.godac.jamstec.go.jp/jedi/player/j/2K0485IN0080Hp04-08 | 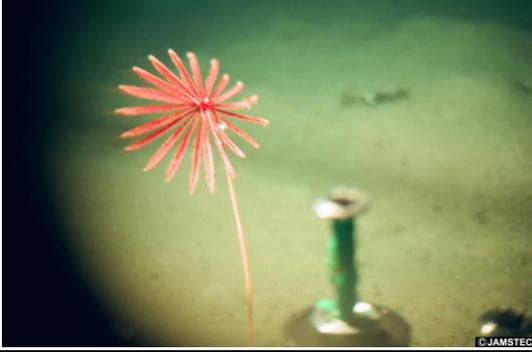 | ムーランルージュ | マバラマキエダミユリ | ヨミチヒロウミユリ | イボウミユリ | A |
| 94 | https://www.godac.jamstec.go.jp/jedi/player/j/HPD0188HDTV0267 | 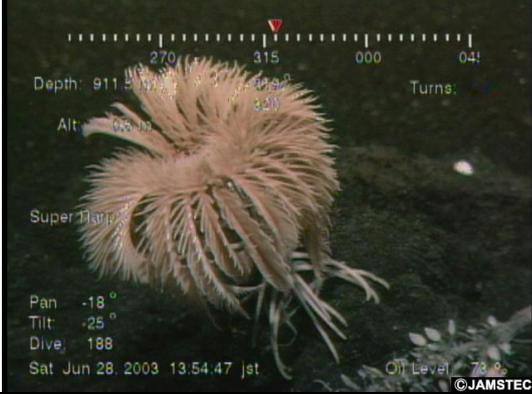 | ムーランルージュ | マバラマキエダミユリ | ヨミチヒロウミユリ | イボウミユリ | B |
| 95 | https://www.godac.jamstec.go.jp/jedi/player/j/2K1107IN0031Hp01-34 | 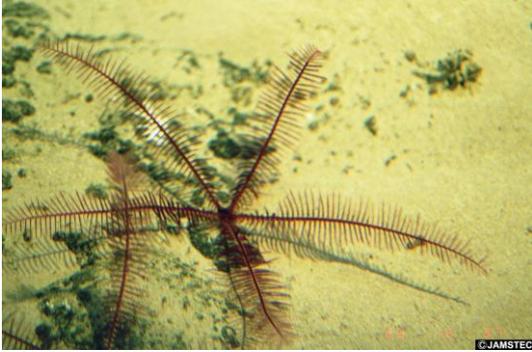 | イツウデウミシダ属 | トフクウミシダ属 | トゲウミシダ属 | イロウミシダ属 | A |
| 96 | https://www.godac.jamstec.go.jp/jedi/player/j/2K0513IN0094Hp03-20 | 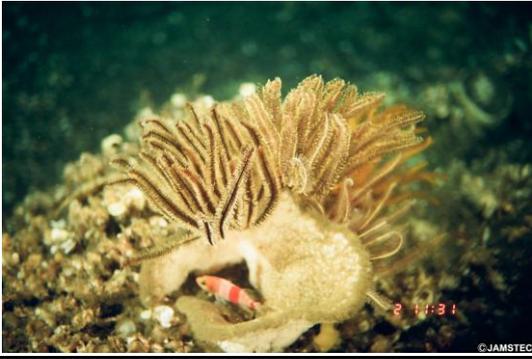 | フトアシウミシダ | キタアシナガヒメウミシダ | ヒゲウミシダ | アシナガヒメウミシダ | A |

Table 1: Table of questions, images and answers (Continue)

| | | | | | | | |
|---|---|---|---|---|---|---|---|
| 97 | https://www.godac.jamstec.go.jp/jedi/player/j/KAIKO0151ASC0237Hp07-21 | 画像の生物は何でしょうか？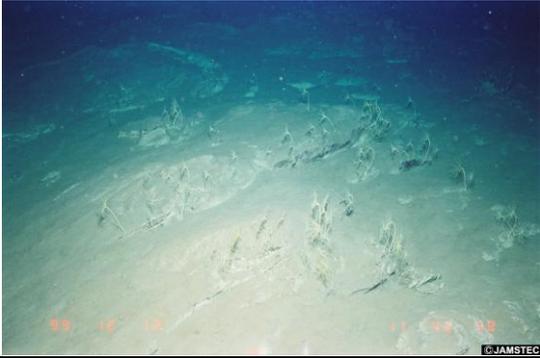 | ムーランルージュ | マバラマキエダウミユリ | ヨミチヒロウミユリ | イボウミユリ | C |
| 98 | https://www.godac.jamstec.go.jp/jedi/player/j/2K0636IN0224Hp07-16 | 画像の生物は何でしょうか？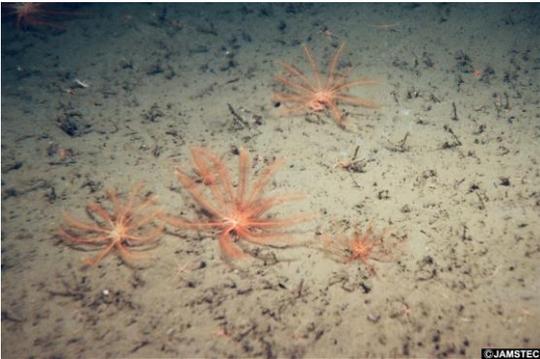 | フトアシウミシダ | キタアシナガヒメウミシダ | ヒゲウミシダ | アシナガヒメウミシダ | B |
| 99 | https://www.godac.jamstec.go.jp/jedi/player/j/3K0385ASC0046Hp02-10 | 画像の生物は何でしょうか？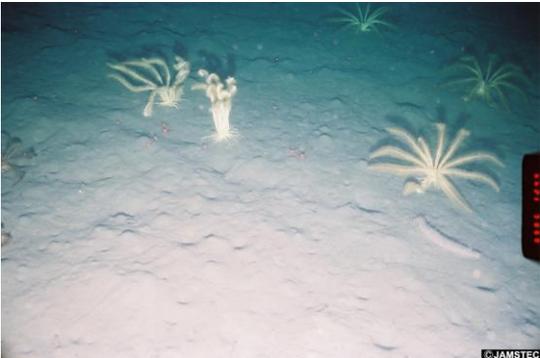 | フトアシウミシダ | キタアシナガヒメウミシダ | ヒゲウミシダ | アシナガヒメウミシダ | C |
| 100 | https://www.godac.jamstec.go.jp/jedi/player/j/2K0343IN0111Hp04-15 | 画像の生物は何でしょうか？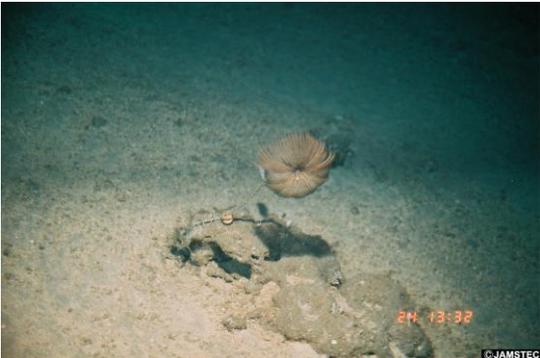 | ムーランルージュ | マバラマキエダウミユリ | ヨミチヒロウミユリ | イボウミユリ | D |